\documentclass[11pt]{article}

\usepackage[final]{acl}
\usepackage{booktabs}
\usepackage{array}

\usepackage{times}
\usepackage{latexsym}
\usepackage{amsmath}
\usepackage[T1]{fontenc}

\usepackage[utf8]{inputenc}

\usepackage{microtype}

\usepackage{inconsolata}
\usepackage{needspace}

\usepackage{graphicx}
\usepackage{subcaption}
\usepackage{stfloats}
\usepackage{float}

\usepackage[most]{tcolorbox}  
\usepackage{xurl}
\usepackage[capitalise]{cleveref}

\newtcolorbox{promptbox}[1]{%
  enhanced,
  breakable,                 
  colback=gray!5,            
  colframe=black,            
  boxrule=0.8pt,             
  arc=2mm,                   
  left=2mm,right=2mm,top=2mm,bottom=2mm, 
  title={#1},                
  fonttitle=\bfseries,       
  coltitle=white,            
  colbacktitle=black         
}

\newtcolorbox{apppromptbox}[1]{%
  enhanced,
  colback=gray!5,
  colframe=black,
  boxrule=0.6pt,
  arc=1.5mm,
  left=1mm,right=1mm,top=0.4mm,bottom=0.4mm,
  boxsep=0.2mm,
  title={#1},
  fonttitle=\bfseries\small,
  coltitle=white,
  colbacktitle=black,
  before skip=1.5pt,
  after skip=1.5pt
}

\title{Sell More, Play Less: Benchmarking LLM Realistic Selling Skill}

\author{%
  \textnormal{Xuanbo Su}\textsuperscript{1},
  \textnormal{Wenhao Hu}\textsuperscript{1},
  \textnormal{Le Zhan}\textsuperscript{1},
  \textnormal{Yuting Xie}\textsuperscript{1},
  \textnormal{Kailin Lyu}\textsuperscript{4,5},
  \textnormal{Kaijie Chen}\textsuperscript{2}, \\
  \textnormal{Ziwei Li}\textsuperscript{3},
  \textnormal{Yeqiang Wang}\textsuperscript{6,7},
  \textnormal{Haibo Su}\textsuperscript{1},
  \textnormal{Yunzhang Chen}\textsuperscript{1},
  \textnormal{Ling Huang}\textsuperscript{1}\thanks{\;Corresponding author.} \\[1pt]
  \footnotesize \textsuperscript{1}Bairong Inc.\quad
  \textsuperscript{2}Tongji University\quad
  \textsuperscript{3}King Abdullah University of Science and Technology \\
  \footnotesize \textsuperscript{4}Institute of Automation, Chinese Academy of Sciences\quad
  \textsuperscript{5}Zhongguancun Academy \\
  \footnotesize \textsuperscript{6}Shanghai Jiao Tong University\quad
  \textsuperscript{7}Shanghai Innovation Institute \\[1pt]
  \footnotesize \texttt{\{xuanbo.su, leo\}@brgroup.com}
}

\begin{document}
\maketitle
\begin{abstract}
Sales dialogues require multi-turn, goal-directed persuasion under asymmetric incentives, which makes them a challenging setting for large language models (LLMs). Yet existing dialogue benchmarks rarely measure deal progression and outcomes. We introduce SalesLLM benchmark, a bilingual (ZH/EN) benchmark derived from realistic applications covering Financial Services and Consumer Goods, built from 30{,}074 scripted configurations and 1{,}805 curated multi-turn scenarios with controllable difficulty and personas. We propose a fully automatic evaluation pipeline that combines (i) an LLM judge for sales-process progress, 
and (ii) fine-tuned BERT classifiers for end-of-dialogue buying intent. To improve simulation fidelity, we train a user model, CustomerLM, with SFT and DPO on 8{,}000+ crowdworker-involved sales conversations, reducing role inversion from 17.44\% (GPT-4o) to 8.8\%. SalesLLM benchmark scores correlate strongly with human ratings (average Pearson $r=0.86$; inter-annotator agreement Krippendorff's $\alpha=0.86$, 500 annotations per model). Across 15 mainstream LLMs, in Chinese the strongest models are competitive with \emph{typical (junior-to-intermediate) human salespeople}---not with sales experts---while weaker ones fall below this baseline, and cross-lingual consistency remains poor. SalesLLM benchmark serves as a scalable benchmark for outcome-oriented sales agents. Our code and data are released at \url{https://github.com/Bairong-Xdynamics/Benchmarking-LLM-Realistic-Selling-Skill}.
\end{abstract}
\section{Introduction}
\begin{figure*}[ht]
    \centering
    \includegraphics[width=\linewidth]{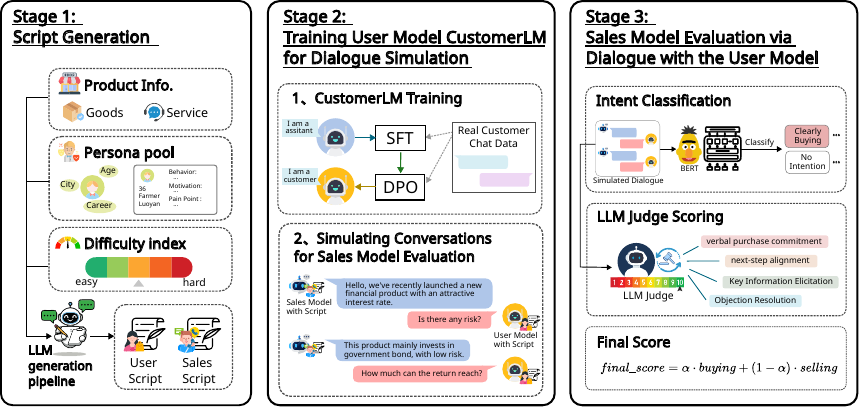}
\caption{
Overview of the SalesLLM benchmark: (1) Standardized Script Generation: Synthesizing role-play scripts by integrating product inventories, persona pools, and difficulty controls. (2) CustomerLM Development \& Simulation: Training a specialized simulator via SFT and DPO to conduct simulated dialogues with evaluated sales LLMs. (3) Evaluation Pipeline: Evaluating sales performance through a dual-metric evaluation framework combining BERT-based intent classification for outcomes and LLM-as-a-judge scoring for process quality.}
    \label{fig:main}
\end{figure*}

Large Language Models (LLMs) are increasingly deployed in goal-directed interactions, with sales emerging as a high-impact application \citep{saleshat, murakhovska-etal-2023-salespeople, chang2024injectingsalespersonsdialoguestrategies, zhang2025aisalesmanreliablelargelanguage, cheng-etal-2025-exploring}. Existing benchmarks, however, measure dialogue quality \citep{liu2016not, sai2020surveyevaluationmetricsused} rather than the outcome-oriented abilities that define selling: proactive influence and conversion. Selling is a distinct form of \textbf{asymmetric persuasion}, in which the agent must shift a counterpart's attitude against possible resistance \citep{Petty1986ElaborationLikelihood, wang2020persuasiongoodpersonalizedpersuasive}. To close this gap, we introduce the \textbf{SalesLLM benchmark} for evaluating proactive persuasion in realistic sales settings. We synthesize 30,074 task scripts and curate 1,805 diverse multi-turn scenarios spanning Financial Services and Consumer Goods. As Figure~\ref{fig:main} illustrates, evaluation is dual-track: an LLM judge rates process quality, while BERT classifiers estimate purchase intention.

For realistic simulation we build \textbf{CustomerLM}, a specialized user simulator trained on 8,000+ crowdworker dialogues with SFT and DPO \citep{ouyang2022traininglanguagemodelsfollow, rafailov2023directpo}, which corrects the formal-language bias and role confusion of general-purpose simulators.

Our contributions are: (1) the SalesLLM benchmark, with 1,805 multi-turn scenarios in Chinese and English; (2) CustomerLM, a realistic user simulator that reduces role inversion; (3) an automated dual-scoring framework for sales evaluation; and (4) a comprehensive evaluation exposing performance gaps in asymmetric persuasion.
\section{Related Work}
\label{rw}

\paragraph{Outcome-oriented multi-turn evaluation.}
Interactive benchmarks have moved beyond single-response quality toward goal-conditioned episodes. Sotopia \citep{zhou2024sotopiainteractiveevaluationsocial} scores social goal completion in open-ended interaction, DailyPersuasion \citep{jin-etal-2024-persuading} supplies large-scale persuasive dialogues across domains, and agent benchmarks such as AgentBoard \citep{agentboard2024} and RoleLLM \citep{wang2024rolellmbenchmarkingelicitingenhancing} add progress tracking and role fidelity. All share a cooperative or neutral counterpart with no incentive to resist. Sales inverts that assumption---the customer may push back---so measuring selling skill requires an explicit conversion objective and calibrated resistance, in the spirit of behavioral test suites \citep{ribeiro-etal-2020-beyond} and difficulty-calibrated agent evaluation \citep{guan2024richelieuselfevolvingllmbasedagents}.

\paragraph{LLMs for sales dialogue.}
Work targeting sales is sparse and mostly agent-side: \citet{murakhovska-etal-2023-salespeople} contrast salespeople with conversational recommenders, \citet{saleshat} deploy an LLM-based social robot for sales conversations, \citet{chang2024injectingsalespersonsdialoguestrategies} inject salesperson strategies via chain-of-thought reasoning, \citet{cheng-etal-2025-exploring} study personality-aware agents, and \citet{zhang2025aisalesmanreliablelargelanguage} pursue reliable telemarketing. These efforts improve the agent but evaluate on small, static datasets, without a reusable protocol that scores whether a deal actually advanced.

\paragraph{User simulation.}
Since the counterpart is simulated, simulator fidelity bounds what an interactive benchmark can measure. Metaphorical simulators \citep{sun-etal-2024-metaphorical} and DuetSim \citep{luo-etal-2024-duetsim} target task-oriented dialogue, USP \citep{wang-etal-2025-know} models implicit user profiles, and UserLM \citep{naous2025flipping} trains dedicated user models; yet reliability studies report persistent drift \citep{sekulic2024reliable}, with simulators answering over-formally and sliding back into the assistant role. A sales customer must also sustain resistance across turns, making role stability the binding constraint---hence CustomerLM.

\paragraph{Automatic scoring and its failure modes.}
Automatic judges are scalable but fragile: style and length inflate verdicts \citep{dubois2025lengthcontrolledalpacaevalsimpleway}, arenas need careful construction to stay discriminative \citep{li2024arenahard}, adversarial framing tricks judges into rewarding incorrect answers \citep{hwang-etal-2025-trick}, and models differ markedly in persuasion susceptibility \citep{bozdag2026persuadecanframeworkevaluating}. We therefore pair an LLM judge---restricted to progression evidenced in the customer's own turns---with a supervised intent classifier, and validate both against human ratings.

\section{SalesLLM Benchmark}
\label{sec:salesllm}

SalesLLM evaluates the selling and persuasive ability of LLMs through three components (Figure~\ref{fig:main}). First, standardized scripts define a scenario for both the salesperson and the customer role. Second, CustomerLM acts as a virtual customer that engages the evaluated sales LLM in multi-turn dialogue. Third, an evaluation pipeline quantifies overall selling proficiency. We describe each component below.

\begin{figure*}[t]
\centering
\includegraphics[width=\linewidth]{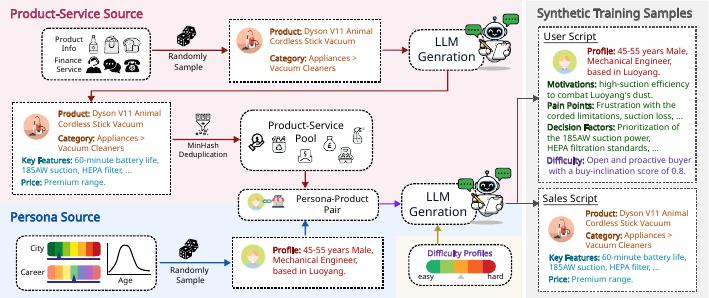}
\caption{The Workflow of the Script Generation Pipeline. The construction process follows a logical transition from raw resources to synthesized scripts:
(1) Raw Component Generation: The pipeline establishes three pillars: Product Information, where raw items are expanded into detailed profiles and deduplicated via MinHash; Persona Generation, involving the sampling of demographic attributes (Age, City, and Career); and the specification of Difficulty Profiles.
(2) Persona-Product Pairing: The processed product-service pool is mapped to the sampled personas to form unique Persona-Product Pairs as the contextual foundation for the scenario.
(3) LLM-based Information Synthesis: Based on these pairs and the assigned difficulty, LLMs generate the final scripts, including a User Script with product-specific motivations and pain points, alongside a Sales Script containing structured product features and selling points.} 
\label{fig:data-generation-pipeline}
\end{figure*}

\subsection{Script Generation}
\label{salesllm:script}
Evaluating sales LLMs first requires high-fidelity scenarios. A realistic scenario pairs a salesperson who commands detailed product knowledge and selling points with a customer who has genuine needs, distinct traits, and a specific purchase propensity. Following established practice \citep{wang2024rolellmbenchmarkingelicitingenhancing,agentboard2024}, we build scripts on three pillars---Product Inventories, Base Personas, and Controllable Difficulty---and synthesize from them concrete scenarios containing a script for each role. Figure~\ref{fig:data-generation-pipeline} visualizes the process.

\paragraph{Product Inventories.}
Our inventory spans two domains: consumer goods and financial services. For consumer goods we seed generation with the Amazon-Reviews-2023 dataset \citep{hou2024bridging}: five product descriptions from one category serve as context for an LLM that synthesizes new profiles with specifications, key features, and selling points. We then apply MinHash deduplication for diversity, yielding 10,074 unique products. For financial services we collect 300 real-world products and run the identical pipeline for cross-domain consistency, producing 20,000 samples.

\paragraph{Base Personas.}
Customer profiles rest on three demographic attributes---age, occupation, and city---with the city attribute coupled to specific decision factors (Appendix~\ref{app:decision-factor-taxonomy}). Sampling these attributes randomly yields a large pool of base personas, i.e., a diverse population of prospects that the next stage enriches with product-conditioned psychology. Figure~\ref{fig:persona-age} (Appendix~\ref{app:age_factor}) shows how decision factors are distributed across age groups.

\paragraph{Controllable difficulty and intent.}
Scenario difficulty---how hard it is for the agent to promote the product---is modulated purely by prompting. We define five calibrated tiers: \emph{easy}, \emph{medium}, \emph{hard}, \emph{very hard}, and \emph{adversarial}. Each tier has its own system prompt encoding a buy propensity and a canonical buyer style, from \emph{cooperative} to \emph{adversarial}; full profiles are in Appendix~\ref{app:difficulty-profiles}. This controllable design allows the same product and base persona to be instantiated under systematically varied conditions, consistent with behavioral test suites \citep{ribeiro-etal-2020-beyond}, user-simulation-based evaluation in task-oriented dialogue \citep{sun-etal-2024-metaphorical,luo-etal-2024-duetsim}, and difficulty-calibrated agent benchmarks \citep{agentboard2024,guan2024richelieuselfevolvingllmbasedagents}.

Explicit difficulty control is necessary because LLMs are demonstrably susceptible to persuasion bias: strategically framed language can push LLM judges to inflate scores for incorrect responses \citep{hwang-etal-2025-trick}, and models differ widely in how readily they are persuaded \citep{bozdag2026persuadecanframeworkevaluating}. Without a calibrated difficulty axis, a simulated customer may capitulate to any argument, inflating scores and masking real capability differences. Controlling buyer resistance---from a cooperative \emph{easy} customer to an adversarial evaluator bent on disqualifying vendors---guards against this bias and spans the full range of real purchase intentions.

To verify that this prompt-only mechanism induces meaningful behavioral variation, we aggregate scores over six representative models (Gemini-3-Flash, Qwen2.5-72B, Doubao-1.5-pro, DeepSeek-Chat, GLM-4.6, and Qwen3-Max). Performance declines monotonically from the \emph{easy} tier (6.79) to the \emph{adversarial} tier (4.98) (Table~\ref{tab:difficulty-scores}), confirming that the tiers are effective.

\paragraph{Design of the propensity knob, and why the score is not circular.}
Difficulty is realized \emph{solely} by writing a prior buy-propensity value $p_k$ and a buyer style into the \emph{customer's} system prompt (Table~\ref{tab:difficulty-profiles}); the salesperson setup, simulator, and scoring pipeline are fixed across tiers. Since one half of the final score is end-of-dialogue buying intent, this raises a circularity concern: that half might merely read back the injected propensity, making Table~\ref{tab:difficulty-scores} tautological. It does not. Across our runs (four models, $\approx$500 dialogues each), dialogues ending in a closed deal (intent class~A) and those that do not have \emph{near-identical} mean injected propensity ($0.33$ vs.\ $0.31$): $p_k$ is a prior on disposition, not a switch on the outcome---a low-propensity customer is genuinely persuadable and a high-propensity one can still be lost. Consistent with this, human agreement holds \emph{within} each tier (mean $r=0.83$; Section~\ref{sec:human-eval}), so the metric is not a between-tier artifact.

\begin{table}[ht]
\centering
\small
\caption{Average SalesLLM benchmark scores by difficulty level, aggregated over Gemini-3-Flash, Qwen2.5-72B, Doubao-1.5-pro, DeepSeek-Chat, GLM-4.6, and Qwen3-Max. Difficulty is controlled solely through the customer's system prompt.}
\label{tab:difficulty-scores}
\begin{tabular}{lcc}
\toprule
\textbf{Difficulty} & \textbf{Avg.\ Score} & \textbf{\# Scenarios} \\
\midrule
\emph{easy}      & 6.79 & 2{,}779 \\
\emph{medium}    & 6.22 & 2{,}998 \\
\emph{hard}      & 5.45 & 1{,}219 \\
\emph{very hard} & 5.33 & 3{,}320 \\
\emph{adversarial} & 4.98 & 2{,}157 \\
\bottomrule
\end{tabular}
\end{table}

\paragraph{Sales Scenario Script Generation.}
We synthesize scenarios by pairing product inventories with base personas. So that simulated customers behave realistically, an LLM enriches each customer script with motivations, pain points, and decision factors, all conditioned on the target product's attributes.

Enrichment at this stage applies only to the customer side; the salesperson script is initialized with detailed specifications and key selling points. The difficulty setting is written directly into the customer script as part of the persona profile.

This process yields 30,074 unique dialogue scripts (examples in Appendix~\ref{app:sc_example}). For evaluation we curate a high-quality subset of 1,000 Chinese scripts and a parallel English version of 805 scripts that survive manual quality filtering.

\subsection{User Model}
\label{salesllm:user_model}
Authentic simulation requires a user model that mirrors real customer behavior. Existing user simulators are typically trained on open-domain or synthetic data~\cite{wang-etal-2025-know, naous2025flipping}, which leaves them poorly suited to playing a prospect in a sales context. Two failure modes recur: \textbf{language bias} and \textbf{role inversion}. Because such models are post-trained to be helpful assistants, they answer at length and in an over-formal register that real customers never use.

\begin{figure}[ht]
    \centering
    \includegraphics[width=\linewidth]{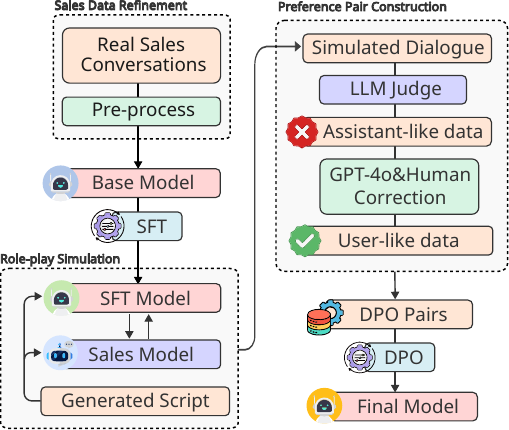}
    \caption{The training pipeline of CustomerLM: (1) SFT on 8,000+ human-annotated sales dialogues to learn fundamental customer behaviors; (2) DPO using 268 contrastive pairs---refined through LLM judging and human correction---to suppress assistant-style bias and role inversion for enhanced persona consistency.}
    \label{fig:user_model}
\end{figure}

We therefore build CustomerLM, a simulator designed to close the gap between assistant-like behavior and realistic customer engagement. As Figure~\ref{fig:user_model} shows, training has two stages: SFT~\citep{ouyang2022traininglanguagemodelsfollow} followed by DPO~\citep{rafailov2023directpo}. We first fine-tune Qwen3-8B \citep{qwen3_2025} on 8,000+ human-annotated sales dialogues to acquire basic customer behavior (statistics in Table~\ref{tab:user_train_data}). For DPO, we generate multi-turn interactions on a held-out validation set between the SFT model and three sales LLMs (GPT-4o, GLM-4.6~\cite{glm4_6_2025}, and Qwen2.5-72B~\cite{qwen2025qwen25technicalreport}). An LLM judge flags degraded user turns---role inversion or language bias---which we repair with GPT-4o and manual inspection to form 268 high-quality contrastive pairs. This targeted preference learning suppresses assistant-style tendencies and keeps the persona consistent across long sales dialogues.

Our ablation study shows that CustomerLM attains markedly better dialogue quality and role consistency than existing simulators; comparisons with GPT-4o, UserLM~\cite{naous2025flipping}, and USP~\cite{wang-etal-2025-know} appear in Section~\ref{exp:ablation_user}.

Detailed training data statistics are provided in Appendix~\ref{app:user-train-data}.

\subsection{Scoring Methodology}
\label{salesllm:dialog_scoring}
We score sales models along two dimensions: \textit{Buying Intent} (outcome-oriented) and \textit{Selling Performance} (process-oriented). Together they balance the immediate conversion result against the proficiency that drives a deal toward closure. Both are computed from the simulated transcripts produced by the interactions of Section~\ref{salesllm:user_model}.

\subsubsection{Scoring User Buying Intention}
We operationalize buying intent by assigning each transcript to one of five levels reflecting latent purchase probability, mapped to $\{2, 4, 6, 8, 10\}$: F (Insulting), X (Perfunctory), C (No Intention), B (Potential Interest), and A (Clear Purchase Intent).

Because buying intent surfaces as fairly explicit signals in sales dialogue, we fine-tune BERT-like models \citep{devlin-etal-2019-bert}---chinese-roberta-wwm-ext \citep{cui-etal-2020-revisiting} for Chinese, roberta-base \citep{liu2019roberta} for English---on 19,178 real-world marketing dialogues in both languages whose intent labels were assigned by human experts. This choice follows evidence \citep{zhang2025bert} that such models capture surface-level patterns well enough to outperform far larger generative LLMs at detecting explicit intent markers in text.

The results bear this out (\cref{tab:bert-comparison}): our fine-tuned BERT is far more accurate than GPT-4o (93.51\% vs.\ 69.6\% in Chinese; 92.94\% vs.\ 68.85\% in English), indicating that domain-specific labels on these architectures give a more reliable read on user intent in complex marketing scenarios (see Appendix~\ref{app:bert-confusion}).

\paragraph{Is the gap prompting or capability?}
The comparison above uses a zero-shot instruction, so part of the gap may reflect prompt design rather than model capability. We therefore re-evaluated GPT-4o with a 10-shot prompt (two worked examples per intent class; Appendix~\ref{app:fewshot-intent}), holding test sets, label set, and decoding fixed. Few-shot prompting helps meaningfully ($+8.8$ points ZH, $+13.1$ EN; Table~\ref{tab:prompt-controlled-intent}), so prompting explains a non-trivial share of the original gap. A sizeable margin nonetheless remains ($\approx$15 points ZH, $\approx$11 EN), so the domain-specialized classifier is still the more reliable intent signal here. We do not claim the residual deficit is intrinsic to GPT-4o (see Limitations).

\begin{table}[ht]
\centering
\small
\setlength{\tabcolsep}{5pt}
\caption{Prompt-controlled buying-intent classification (accuracy, \%). Few-shot prompting helps GPT-4o substantially, yet a sizeable gap remains.}
\label{tab:prompt-controlled-intent}
\begin{tabular}{lcc}
\toprule
\textbf{Intent classifier} & \textbf{ZH} & \textbf{EN} \\
\midrule
GPT-4o, zero-shot & 69.60 & 68.85 \\
GPT-4o, 10-shot & 78.42 & 81.92 \\
Fine-tuned BERT (ours) & \textbf{93.51} & \textbf{92.94} \\
\bottomrule
\end{tabular}
\end{table}

\subsubsection{Scoring Salespeople Selling Performance}
Drawing on sales psychology, we assess tactical proficiency along four dimensions: (1) verbal purchase commitment, (2) concrete next-step alignment, (3) key information elicitation, and (4) objection resolution effectiveness. An LLM-as-a-judge assigns $S_{sell} \in [0, 10]$, reflecting how far the dialogue advanced and how skillfully the techniques were applied. Concretely, \emph{verbal purchase commitment} credits the customer's own expression of willingness to buy; \emph{concrete next-step alignment} requires a specific, mutually acknowledged next action rather than a vague ``I'll think about it''; \emph{key information elicitation} asks whether the agent obtained the facts needed to qualify and tailor the offer (budget, horizon, risk tolerance, competing options, decision authority); and \emph{objection resolution effectiveness} credits an objection only when specific evidence is given \emph{and} the customer's later turns show the concern has eased. Credit always depends on customer-side evidence, never on the agent's assertions alone. Full definitions and the annotation protocol are in Appendix~\ref{app:subdim}, and each sub-dimension is validated separately against human annotations in Section~\ref{sec:human-eval}.

\paragraph{Mitigating known LLM-judge biases.}
Position, verbosity, and persuasion biases are mitigated by construction: the judge scores each dialogue independently on an absolute 0--10 rubric (no pairwise comparison, hence no ordering to swap), and credits only deal progression evidenced in the \emph{customer's} turns, ignoring assistant claims the customer did not accept. Appendix~\ref{app:judge-bias} details each mechanism.

\subsubsection{Final Score}
The final score is a weighted combination of the buying-intent score ($S_{buy}$) and the selling-performance score ($S_{sell}$), given in Equation~\eqref{eq:1}:

\begin{equation}
\text{Final\_Score} = \alpha \cdot S_{buy} + (1 - \alpha) \cdot S_{sell}
    \label{eq:1}
\end{equation}

We set $\alpha = 0.5$ throughout, weighting conversion result and process quality equally; a sensitivity analysis over $\alpha \in \{0.1, 0.3, 0.5, 0.7\}$ confirms that $\alpha = 0.5$ maximizes agreement with human scores (Appendix~\ref{app:alpha}). Agreement with human scores peaks at $\alpha=0.5$ for three of four models and in the average (Appendix~\ref{app:alpha}), so neither dimension alone accounts for human judgment. This design avoids the single-aspect focus common to existing benchmarks \citep{dubois2025lengthcontrolledalpacaevalsimpleway}.

\subsubsection{Human Score Justification}
To verify reliability, eight annotators independently scored 2,000 dialogues (500 for each of four representative models) on the same 0--10 scale. These dialogues are disjoint from everything used to build or tune the pipeline (BERT training data, judge-prompt development), so the correlation is measured on held-out rather than fitted data. As Section~\ref{sec:human-eval} reports, the automated scores align strongly with human judgment (average Pearson $r = 0.86$); the agreement holds within every difficulty tier and for each selling sub-dimension, and inter-annotator agreement is substantial (Krippendorff's $\alpha = 0.86$). Together these support the framework for large-scale automated assessment.

\section{LLMs Results on SalesLLM Benchmark}
\label{sec:results}

We now evaluate mainstream LLMs on SalesLLM. Models are run in both Chinese and English to probe cross-lingual selling ability, and we compare automated scores against human ratings to validate the methodology.

\subsection{Experimental Setup}
We evaluate 15 mainstream LLMs spanning multiple model sizes across Chinese and English scenarios using two user simulators: GPT-4o \cite{openai2024gpt4ocard} and our trained user model. For models accessed via official APIs, we test Doubao-1.5pro-32k \cite{doubao32k2025}, Qwen3-max \cite{qwen3_2025}, Deepseek-chat \cite{deepseekchat2025}, GLM-4.6 \cite{glm4_6_2025}, GPT-4o, GPT-5-nano \cite{openai2025gpt5}, Gemini-3-series \cite{deepmind2025gemini}, and Xiaomi-MiMo-V2 \cite{mimo2025flash}.
For open-weights models, we evaluate Llama-3.3-70B \cite{grattafiori2024llama3herdmodels}, Qwen(Qwen3-8B/32B, Qwen2.5-72B) \cite{qwen3_2025,qwen2025qwen25technicalreport}, and Gemma-3-27B \cite{gemmateam2025gemma3technicalreport}.
All models use the same sampling parameters (temperature 0.8, top\_p 0.99, max\_tokens 2048) and at most 20 dialogue rounds. Proprietary models are accessed through official API endpoints, open-weights models through vLLM~\cite{kwon2023efficient}. For the human baseline, we built the interface in Figure~\ref{fig:salesinterface}, through which human salespeople sell to a customer played by GPT-4o. These salespeople were recruited from the financial and general consumer-product domains with at least one year of experience, and represent \textbf{typical/average human performance}, not top sales experts.
\subsection{Overall Performance}
\label{exp:over_p}
\begin{figure*}[t]
    \centering
    \begin{minipage}{0.42\textwidth}
        \centering
        \includegraphics[width=\linewidth]{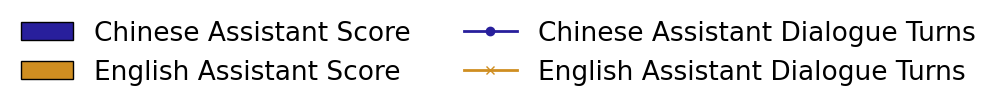}
    \end{minipage}
    \vspace{0.5em}

    \begin{subfigure}{0.49\textwidth}
        \centering
        \includegraphics[width=\linewidth]{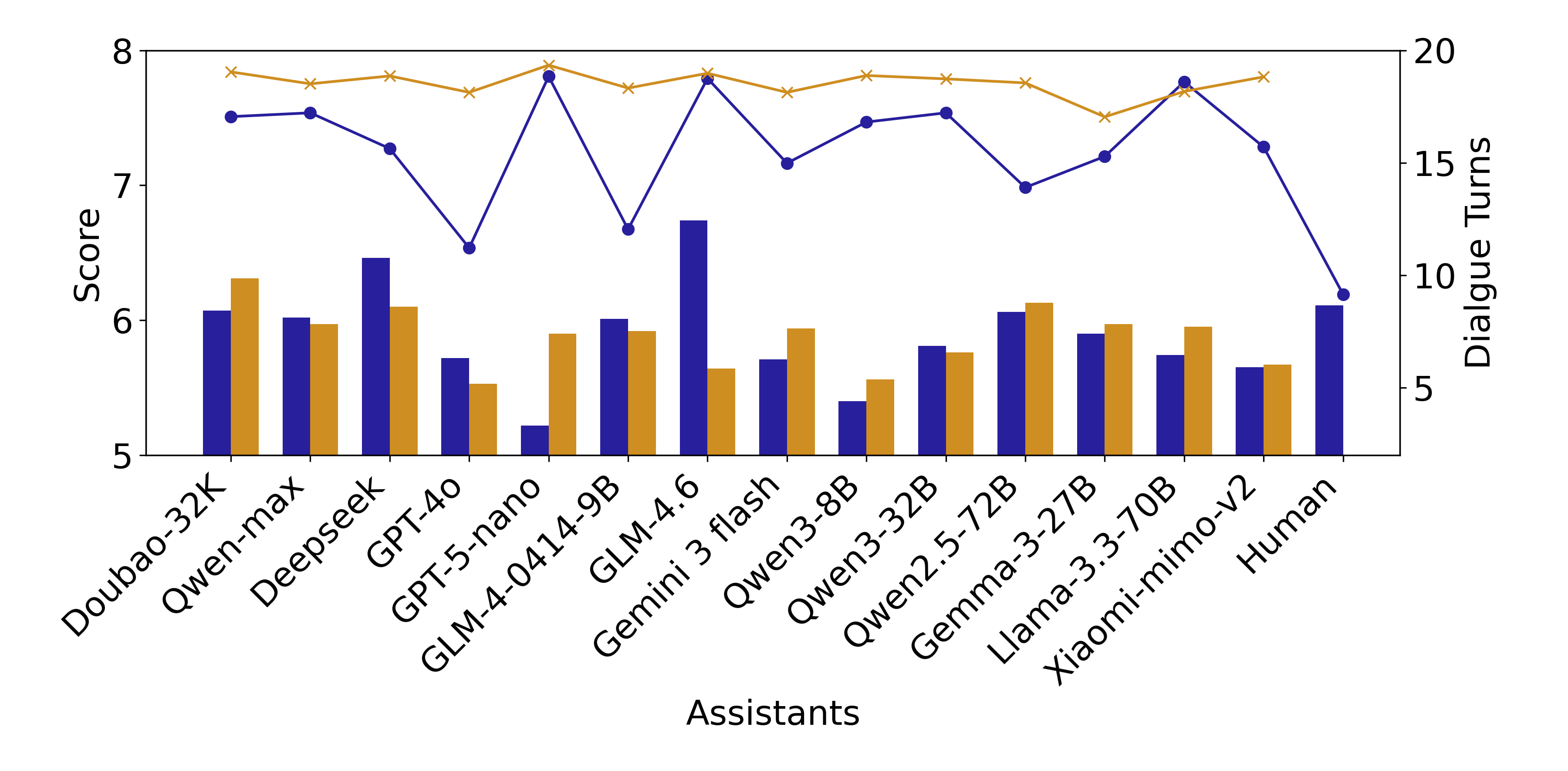}
        \caption{GPT-4o as User Model}
        \label{fig:score-a}
    \end{subfigure}
    \hfill
    \begin{subfigure}{0.49\textwidth}
        \centering
        \includegraphics[width=\linewidth]{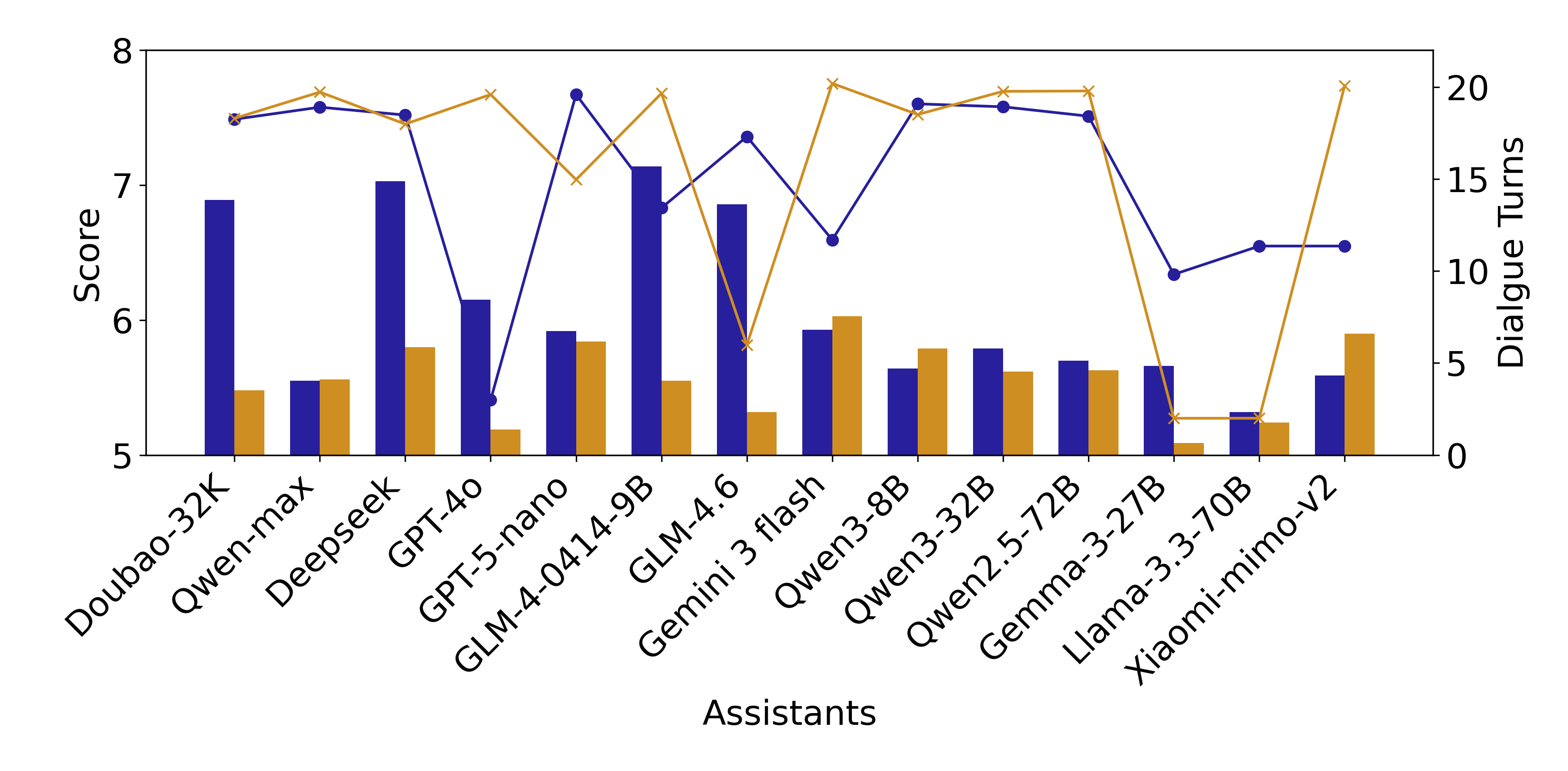}
        \caption{CustomerLM as User Model}
        \label{fig:score-b}
    \end{subfigure}

    \caption{Distribution of SalesLLM benchmark scores vs. dialogue turns across 1,805 scenarios (Chinese and English).}
    \label{fig:score}
\end{figure*}

Table~\ref{tab:overall_results} and Figure~\ref{fig:score} report overall performance. Selling ability varies widely, and stronger models score consistently higher. Notably, top-tier models such as Doubao-1.5 and GLM-4.6 exceed our human baseline (Human Salespeople Group) in Chinese, indicating real potential for LLMs in sales. We stress that this baseline consists of practitioners with at least one year of experience---i.e., \emph{junior-to-intermediate} practitioners---so these results indicate parity with typical human salespeople and should not be read as parity with sales experts; an expert-calibrated baseline is left to future work (see Limitations).

Models that excel in Chinese do not necessarily transfer: Doubao-1.5 drops sharply in English against CustomerLM (6.89 $\rightarrow$ 5.48), whereas Gemini-3 stays consistent across languages (6.39/6.03). Qualitatively (Appendix~\ref{app:case_study}), strong models such as DeepSeek-Chat act proactively---asking closing questions and driving conversion---while weaker ones behave like passive Q\&A bots.

Table~\ref{tab:overall_results} and Figure~\ref{fig:score} also show that the same model typically averages a different number of turns in Chinese than in English, suggesting that language shapes the sales process itself. We leave a detailed study of this effect to future work.

\paragraph{The simulator changes the conclusion.}
Since any such benchmark scores models against one simulated customer, that choice deserves scrutiny. Across the 15 assistants, per-model scores under the two simulators correlate only moderately in Chinese (Pearson $r=0.57$) and weakly in English ($r=0.24$), and the leaderboard reorders: DeepSeek-Chat and GLM-4.6 tie for first in Chinese under the GPT-4o user, whereas under CustomerLM the top three are GLM-4-9B (7.14), DeepSeek-Chat (7.03), and Doubao-1.5 (6.89), and Gemini-3-pro falls from 6.52 to 5.93. GPT-4o is easier to persuade and compresses differences at the top; CustomerLM resists longer and separates models more sharply. Rankings from a general-purpose simulator should thus be read with caution---why we release CustomerLM with the benchmark.

\begin{table}[h]
\centering
\small
\caption{Overall performance (SalesLLM benchmark score) of different Assistant Models when evaluated against two user simulators: GPT-4o and our CustomerLM Model. Results are reported for both Chinese (ZH) and English (EN) scenarios. The best result is bolded, the second best is underlined, and the third best is marked with $^{\dag}$.}
\label{tab:overall_results}
\begin{tabular}{lcccc}
\toprule
& \multicolumn{2}{c}{\textbf{GPT-4o User}} & \multicolumn{2}{c}{\textbf{CustomerLM}} \\
\cmidrule(lr){2-3} \cmidrule(lr){4-5}
\textbf{Assistant Model} & \textbf{ZH} & \textbf{EN} & \textbf{ZH} & \textbf{EN} \\
\midrule
Doubao-1.5-pro-32k & 6.50 & \underline{6.31} & 6.89$^{\dag}$ & 5.48 \\
Qwen3-Max & 6.48 & 5.97 & 5.55 & 5.56 \\
DeepSeek-Chat (V3.1) & \textbf{6.74} & 6.10 & \underline{7.03} & 5.80 \\
GPT-4o & 6.16 & 5.53 & 6.15 & 5.19 \\
GLM-4-9B & 6.01 & 5.92 & \textbf{7.14} & 5.55 \\
GLM-4.6 & \textbf{6.74} & 5.64 & 6.86 & 5.32 \\
Gemini-3-pro & 6.52$^{\dag}$ & \textbf{6.39} & 5.93 & \textbf{6.03} \\
Qwen3-8B & 5.40 & 5.56 & 5.64 & 5.79 \\
Qwen3-32B & 5.81 & 5.76 & 5.79 & 5.62 \\
Qwen2.5-72B & 6.06 & 6.13$^{\dag}$ & 5.70 & 5.63 \\
Gemma-3-27B & 5.90 & 5.97 & 5.66 & 5.09 \\
Llama-3.3-70B & 5.74 & 5.95 & 5.32 & 5.24 \\
GPT-5-nano & 5.22 & 5.90 & 5.92 & 5.84 \\
Xiaomi-MiMo-V2 & 5.65 & 5.67 & 5.59 & 5.90$^{\dag}$ \\
Gemini-3-flash & 5.71 & 5.94 & 5.93 & \textbf{6.03} \\
\midrule
Human Group & 6.33 & -- & -- & -- \\
\bottomrule
\end{tabular}
\end{table}

\subsection{Human Evaluation Validation}
\label{sec:human-eval}
\begin{figure}[t]
    \centering
    \includegraphics[width=\linewidth]{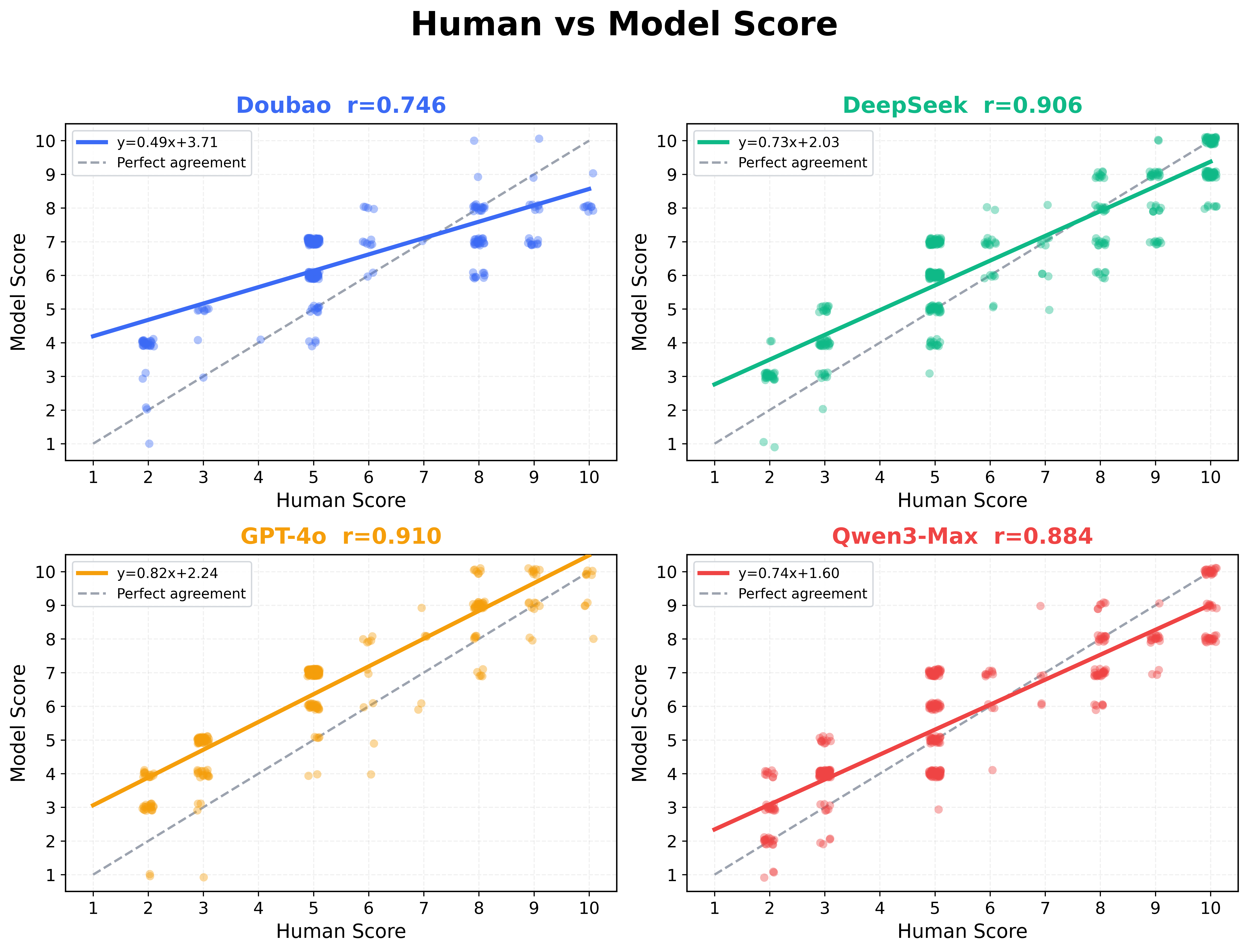}
    \caption{Correlation analysis between Human evaluation scores and SalesLLM benchmark scores. We evaluated four representative models, with 500 dialogues per model annotated by human experts. The overall average Pearson correlation coefficient is $r=0.86$, demonstrating strong alignment between automated and human judgments.}
    \label{fig:plot_1}
\end{figure}

To validate the automated metric, we compare SalesLLM scores with ratings from eight human scorers on 2,000 dialogues (500 randomly sampled for each of four representative models: Doubao, DeepSeek, GPT-4o, and Qwen3-Max), collected through the interface in Figure~\ref{fig:scoreinterface}. Figure~\ref{fig:plot_1} shows strong alignment: the average Pearson correlation is $r=0.86$~\cite{pearson1895note}, well above the conventional threshold for high correlation ($r>0.6$)~\cite{cohen1988statistical}. The annotated dialogues are held out from all pipeline construction and tuning (Section~\ref{salesllm:dialog_scoring}). We report three further controls below: inter-annotator agreement, within-tier agreement, and per-sub-dimension validation; the annotation protocol is in Appendix~\ref{app:subdim}.

\paragraph{Reliability of the human reference.}
Since the reference signal is human, its own reliability must be quantified. On the same 2{,}000-dialogue study, agreement among the eight annotators is substantial on all three standard measures: mean pairwise Pearson $r=0.82$, $\mathrm{ICC}(2,1)=0.82$, and Krippendorff's $\alpha=0.86$ (interval), above the conventional $\alpha \ge 0.80$ threshold (Appendix~\ref{app:reliability}). The ratings underpinning $r=0.86$ are therefore themselves reliable.

\paragraph{Within-tier agreement.}
A correlation pooled across five tiers---whose scores are designed to differ widely---could be inflated by between-tier variance. Recomputing agreement separately within each tier removes that component and gives a mean within-tier $r=0.83$ (range $0.75$--$0.88$; Appendix~\ref{app:reliability}), close to the pooled per-model average of $0.86$. The agreement is therefore not a between-tier artifact: the score tracks human judgment even at fixed difficulty.

\paragraph{Per-sub-dimension validation of the LLM judge.}
The correlations above validate the combined score, but not the LLM-judge dimension on its own. We therefore collected separate human annotations for each of the four Selling sub-dimensions (Section~\ref{salesllm:dialog_scoring}) and correlated each with its automated sub-score: verbal purchase commitment $r=0.86$, concrete next-step alignment $0.79$, key information elicitation $0.80$, and objection resolution effectiveness $0.82$. Every sub-dimension tracks human judgment independently (all $r \ge 0.79$), validating the selling-performance dimension in isolation. Definitions and protocol are in Appendix~\ref{app:subdim}; Appendix~\ref{app:reliability} tabulates all three controls.

\subsection{Multi-Product and Long-Horizon Selling}
\label{sec:multi-long}
To better reflect real sales work, we add two settings: \emph{multi-product selling}, where the assistant must navigate several offerings, and \emph{long-horizon selling}, where it must persist through an initial rejection across multiple touchpoints.

\paragraph{Long-horizon selling.}
Real campaigns often demand sustained engagement after a customer says no. We therefore build long-horizon scenarios from single-session trajectories in the main benchmark that end with clear non-purchase intent, i.e., with no progression toward a purchase.

We evaluate the leaders from our main leaderboard: Doubao-1.5-pro-32k, DeepSeek-Chat (V3.1), and Gemini-3-pro. Each scenario permits up to two follow-up rounds, with the full conversation history carried into the system prompt between rounds; interaction stops early if intent is clearly negative. The final score covers the whole dialogue, from the early conversion attempt to the last.

\paragraph{Multi-product selling.}
Here sales assistants must manage a portfolio of six items per scenario, manually curated to align with each customer persona's likely preferences; the assistant receives concatenated product descriptions generated with our script framework (Section~\ref{salesllm:script}). This assesses two capabilities: (i) dynamic product selection as customer needs evolve, and (ii) bundling strategies beyond single-item promotion. We evaluate the same three top models (Table~\ref{tab:multi-p}).
\begin{table}[h]
\centering
\small
\caption{Performance of multi-product selling of different Assistant Models when evaluated against two user simulators: GPT-4o and our CustomerLM Model. Results are reported for both Chinese (ZH) and English (EN) scenarios.}
\label{tab:multi-p}
\begin{tabular}{lcccc}
\toprule
& \multicolumn{2}{c}{\textbf{GPT-4o User}} & \multicolumn{2}{c}{\textbf{CustomerLM}} \\
\cmidrule(lr){2-3} \cmidrule(lr){4-5}
\textbf{Assistant Model} & \textbf{ZH} & \textbf{EN} & \textbf{ZH} & \textbf{EN} \\
\midrule
Doubao-1.5-pro-32k & \textbf{7.46} & 6.62 & 6.16 & 5.93 \\
DeepSeek-Chat (V3.1) & 6.47 & \textbf{6.96} & 6.24 & 5.67 \\
Gemini-3-pro & 7.17 & 7.03 & \textbf{6.56} & \textbf{5.95} \\
\bottomrule
\end{tabular}
\end{table}

Table~\ref{tab:multi-p} shows that Gemini-3-pro performs strongly across most conditions, leading in English under both user simulators (GPT-4o: 7.03; CustomerLM: 5.95) and dominating in Chinese under CustomerLM (6.56).

\paragraph{Scaling up the portfolio: a 15-SKU study.}
Six hand-curated products is a modest portfolio, so we repeated the evaluation with \textbf{15 SKUs} per scenario (1 target plus 14 competing peers from the same catalog), which sharply increases the burden of dynamic selection and creates real opportunities for bundling. Over 200 scenarios per language with CustomerLM, the same three models score 6.47/6.11 (Doubao-1.5-pro-32k), 6.35/5.65 (DeepSeek-Chat), and 6.54/5.98 (Gemini-3-pro) for ZH/EN: scores hold or slightly improve over the 6-SKU setting---a larger portfolio offers more ways to match the customer---with the ordering preserved, so the finding is not an artifact of portfolio size.

\begin{table}[h]
\centering
\small
\caption{Performance of long horizon selling of different Assistant Models when evaluated against two user simulators: GPT-4o and our CustomerLM Model. Results are reported for both Chinese (ZH) and English (EN) scenarios.}
\label{tab:lh}
\begin{tabular}{lcccc}
\toprule
& \multicolumn{2}{c}{\textbf{GPT-4o User}} & \multicolumn{2}{c}{\textbf{CustomerLM}} \\
\cmidrule(lr){2-3} \cmidrule(lr){4-5}
\textbf{Assistant Model} & \textbf{ZH} & \textbf{EN} & \textbf{ZH} & \textbf{EN} \\
\midrule
Doubao-1.5-pro-32k & \textbf{4.91} & \textbf{4.62} & 4.58 & 4.4 \\
DeepSeek-Chat (V3.1) & 4.58 & 4.48 & 5.92 & \textbf{5.82}\\
Gemini-3-pro & 4.43 & 4.58 & \textbf{6.20} & 4.88 \\
\bottomrule
\end{tabular}
\end{table}

Table~\ref{tab:lh} shows more nuanced results for long-horizon selling, varying markedly across both assistant models and user simulators. Notably, several models recover better against CustomerLM than against GPT-4o, indicating that follow-up persuasion effectiveness is highly sensitive to how the simulator models customer resistance and receptivity to re-engagement---so the choice of user simulator substantially influences any assessment of long-term sales persistence.

All long-horizon scores fall below the single-session results of Table~\ref{tab:overall_results}, as expected when only already-rejected trajectories are selected. More informative is the direction of change: persistence is not uniformly rewarded. For Doubao-1.5-pro in English under CustomerLM, a second approach \emph{lowers} the score against its own first session ($\Delta=-0.34$). The transcript (Appendix~\ref{app:case_c}) shows why---the agent stacks a limited-time discount, social proof, and scarcity framing onto a customer who has just asked for time, and the customer escalates from polite deferral (C) to explicit refusal (X). GPT-4o absorbs such pressure politely, so the cost is invisible there: measuring re-engagement needs a simulator that can become \emph{less} receptive.

\subsection{Ablation Study of CustomerLM Model}
\label{exp:ablation_user}
To validate CustomerLM, we compare it against GPT-4o, UserLM~\cite{naous2025flipping}, and USP~\cite{wang-etal-2025-know} on two held-out test sets: 118 Chinese and 150 English conversations collected via interactions with 10 human contractors (outsourced sales agents). Summary statistics are provided in Appendix Table~\ref{tab:ablation_test_stats}.  

We evaluate each user model by predicting customer turns conditioned on the ground-truth preceding context and comparing them with the actual customer utterances. We report \textbf{BLEU-4} and \textbf{ROUGE-L} \cite{papineni-etal-2002-bleu,lin-2004-rouge} for lexical overlap and \textbf{embedding cosine similarity (Sem.\ Sim.)} \cite{chen2025m3embeddingmultilingualitymultifunctionalitymultigranularity} for semantic alignment. For behavioral authenticity we additionally measure the \textbf{Role Inversion Rate}: the percentage of generated turns in which the model wrongly adopts the salesperson persona (e.g., pitching products), as judged by GPT-4o. Implementation details and summary statistics are in Appendix~\ref{app:user-eval}.

\begin{table}[H]
    \centering
    \small
    \caption{Ablation study comparing CustomerLM with GPT-4o, UserLM, and USP as user models. R-L denotes ROUGE-L and RI denotes role inversion; higher is better except for RI.}
    \label{tab:user_model_ablation}
    \setlength{\tabcolsep}{2pt}
    \begin{tabular*}{\columnwidth}{@{\extracolsep{\fill}}lcccc@{}}
        \toprule
        \textbf{Model} & \textbf{BLEU} & \textbf{R-L} & \textbf{Sem.} & \textbf{RI (\%)} \\
        \midrule
        GPT-4o & 0.10 & 0.07 & 0.57 & 17.44 \\
        UserLM & 0.06 & 0.06 & 0.50 & 21.55 \\
        USP & 0.08 & 0.08 & 0.52 & 18.76 \\
        CustomerLM & \textbf{0.12} & \textbf{0.10} & \textbf{0.59} & \textbf{8.8} \\
        \bottomrule
    \end{tabular*}
\end{table}

\paragraph{Results.} \cref{tab:user_model_ablation} shows that CustomerLM leads on turn-level similarity and cuts role inversion to 8.8\%, i.e., more faithful customer-style behavior. UserLM and USP invert roles more often (21.55\% and 18.76\%) than the generic GPT-4o baseline (17.44\%), and occasionally leak script instructions or slip into a helpful-assistant persona---evidence of how hard a consistent user role is to maintain in sales dialogue. 

\section{Conclusion}

We presented SalesLLM, a benchmark for LLM selling skill with 1,805 difficulty-calibrated scenarios from 30,074 scripts, together with CustomerLM, a user simulator that reduces role inversion, and a dual-metric framework validated against human judgment ($r=0.86$, with substantial inter-annotator and within-tier agreement and per-sub-dimension checks). Top-tier LLMs match \emph{typical (junior-to-intermediate) human salespeople} in Chinese---not experts---but cross-lingual consistency and complex persuasion remain weak, and simulator choice shifts conclusions.

\section*{Limitations}

We identify several primary limitations that constrain the generalizability of our findings. First, our human salesperson baseline consists of practitioners with at least one year of experience, representing entry-level to intermediate rather than expert-level performance. While this provides a reasonable comparison point, LLMs outperforming this baseline only demonstrates they have surpassed junior-to-intermediate capability, not true expertise, and every statement of human parity in this paper should be read in that narrower sense. Recruiting and de-identifying certified senior-salesperson dialogues at benchmark scale was beyond the scope of this work; we also note that, to our knowledge, no public benchmark measures outcome-oriented multi-turn selling skill against calibrated expert performance (existing resources emphasize dialogue quality rather than deal progression), which is part of our motivation. Future work should establish benchmarks against seasoned professionals with 5+ years of experience.

Second, while we employ human calibration to enhance realism, our LLM-based user simulator may not fully capture the complexity of real customer behavior. Customers exhibit unpredictable emotional fluctuations, varying levels of product knowledge, cultural nuances, and complex decision-making processes that are difficult to model computationally \citep{10.1007/978-3-031-54975-5_5,sekulic2024reliable,wang-etal-2025-know}. Additionally, trust building often depends on subtle non-verbal cues that our text-based simulation cannot adequately represent \citep{sun-etal-2024-metaphorical,luo-etal-2024-duetsim}.

Third, although our benchmark supports multi-turn dialogues within a session and includes extended evaluation settings such as long-horizon follow-ups and multi-product selling (Section~\ref{sec:multi-long}), real-world enterprise pipelines often unfold across calendar-time gaps---multiple touches separated by days or weeks, coordinated across channels and CRM systems. SalesLLM benchmark does not yet instantiate these slower-cycle, organization-wide dynamics.

Fourth, we observe that sales agents occasionally hallucinate unauthorized concessions---such as offering discounts not present in their product scripts---to maximize conversion rates, despite explicit instructions to adhere to provided guidelines. This unfaithful persuasion behavior poses challenges for reliable evaluation and echoes broader validity concerns in LLM benchmarks where models may achieve high scores through undesirable shortcuts \citep{li2024arenahard}.

Fifth, our prompt-controlled comparison for intent classification (Section~\ref{salesllm:dialog_scoring}) shows that few-shot prompting recovers a substantial part of GPT-4o's deficit, so we do not claim the residual gap is intrinsic to GPT-4o; a better prompt or a stronger generative model could narrow it further. Our claim is the narrower, pipeline-level one that a domain-specialized classifier is currently the more reliable intent signal at our scale and cost. Relatedly, although our judge avoids pairwise protocols (in which position bias arises) and anchors its score on customer-side progression rather than assistant rhetoric (Appendix~\ref{app:judge-bias}), it remains an LLM with possible residual sensitivity to surface form, and scores come from a single run; human validation (Section~\ref{sec:human-eval}) bounds but does not eliminate this risk.

Finally, part of our data derives from real customer interactions and is subject to privacy and compliance review, so the training corpora for CustomerLM and the intent classifiers are released only as held-out test splits (see Appendix~\ref{app:availability}). This limits reproducibility of the training stages, though not of the reported evaluation numbers.

\section*{Acknowledgements}

We thank Xiaoliang Ma, Juan Du, Qiao Huang, and their colleagues in the Marketing and Wealth Management departments at Bairong Inc.\ for serving as domain experts throughout this work: they contributed practitioner sales expertise, took part in data collection and annotation, and helped ensure that our scenarios and scoring rubric reflect real-world selling practice. We are also grateful to Siji Wan for valuable discussions and insights that shaped the design of this benchmark, and to Prof.\ Ruihua Song of Renmin University of China for her advice and guidance on this work.

\bibliography{custom}
\clearpage

\appendix
\numberwithin{table}{section}
\numberwithin{figure}{section}
\setlength{\textfloatsep}{6pt plus 2pt minus 2pt}
\setlength{\intextsep}{6pt plus 2pt minus 2pt}
\setlength{\floatsep}{6pt plus 2pt minus 2pt}
\setlength{\abovecaptionskip}{4pt}
\setlength{\belowcaptionskip}{2pt}
\renewcommand{\arraystretch}{0.95}

\section{Data and Code Availability}
\label{app:availability}

The 1{,}805 evaluation scripts, the evaluation code, the model rollouts and automatic scoring outputs, and the 2{,}000 human ratings behind our reliability analysis are released in the repository linked in the abstract. The CustomerLM simulator and the fine-tuned bilingual buying-intent classifiers are released as model checkpoints at \url{https://huggingface.co/MultiSense/CustomerLM} and \url{https://huggingface.co/MultiSense/SaleIntent_bert}. The CustomerLM SFT dialogues and the 19{,}178 annotated dialogues used to train the classifiers derive from real customer interactions and cannot be released in full under our data-governance constraints; we release the held-out \emph{test} splits for both (user-likeness; ZH/EN intent), which suffice to reproduce the reported numbers, with the DPO preference pairs to follow after de-identification review. All released data are de-identified.

\section{User-Model Evaluation Details}
\label{app:user-eval}

This section details the automatic evaluation protocol for the user simulator described in \cref{exp:ablation_user}.

\subsection{Evaluation Methodology}
\paragraph{User Likeness (Turn-Level Similarity).}
We evaluate the user model by predicting customer turns conditioned on ground-truth history from crowd worker dialogues. Given a dialogue context (system message + previous turns), we generate the next customer utterance and compute its similarity to the human ground truth using BLEU-4, ROUGE-1/2/L, and embedding cosine similarity (via `bge-m3'). We apply language-specific tokenization and exclude short acknowledgments (length $\le$ 5 chars) from evaluation to ensure meaningful comparison. 
\begin{table}[ht]
\small
    \centering
    \begin{tabular}{cccc}
        \toprule
         \textbf{Product}& \# \textbf{Dialogues} & \textbf{Avg. Turns} & \textbf{Avg. Msg Len} \\
         \midrule
         Security& 189 & 3.32 & 14.07 \\
         Banking & 50 & 4.79 & 10.2 \\
         Fund & 29 & 2.61 & 16.23 \\
         \bottomrule
    \end{tabular}
    \caption{The statistical information of CustomerLM test data. \# Dialogues is the number of dialogues, Avg. Turns is the average turns of the dialogues, Avg. Msg Len is the average user message lengths.}
    \label{tab:user_test_data}
\end{table}

\subsection{Training Data Statistics}
\label{app:user-train-data}
Training data statistics are provided in Table~\ref{tab:user_train_data}.
\begin{table}[ht]
\small
    \centering
    \begin{tabular}{cccc}
        \toprule
         \textbf{Product}& \# \textbf{Dialogues} & \textbf{Avg. Turns} & \textbf{Avg. Msg Len} \\
         \midrule
         Security& 4748 & 2.3 & 16.25 \\
         Banking & 1742 & 5.98 & 12.03 \\
         Fund & 1794 & 5.42 & 11.97 \\
         \bottomrule
    \end{tabular}
    \caption{The statistical information of CustomerLM training data. \# Dialogues is the number of dialogues, Avg. Turns is the average turns of the dialogues, Avg. Msg Len is the average number of words per user message.}
    \label{tab:user_train_data}
\end{table}

\begin{table}[t]
\centering
\small
\setlength{\tabcolsep}{3pt}
\begin{tabular}{lrrr}
\toprule
\textbf{Lang} & \textbf{\#Dia} & \textbf{Avg msgs/conv} & \textbf{Avg user msgs} \\
\midrule
ZH (118\_zh) & 118 & 7.98 & 3.85 \\
EN (150\_en) & 150 & 8.13 & 3.83 \\
\bottomrule
\end{tabular}
\caption{Held-out ablation test set statistics (Sec.~\ref{exp:ablation_user}). Averages exclude the system prompt. Lang is language, Dia is dialogues.}
\label{tab:ablation_test_stats}
\end{table}

Below are the system prompt used with CustomerLM as the user model, along with one example of test data.

\needspace{22\baselineskip}
\noindent
\begin{apppromptbox}{System Prompt}
\small
\textbf{Your task}: You are a customer management expert. The customer is a rep from xx Securities promoting a stable return of 0.035 (50,000 yuan over 3 years = 5,250 yuan, bank-supervised). Suggests adding the company's WeChat for assistance.\\[1pt]
\textbf{Rules}: (1) Use English strictly. (2) Be realistic; do not mention impossible things. (3) Avoid personal life or off-topic content. (4) Use conversational language; keep responses brief. (5) Do not fabricate facts; respond from a customer perspective. (6) No parenthetical actions or inner thoughts.
\end{apppromptbox}

\needspace{18\baselineskip}
\noindent
\begin{apppromptbox}{One example of test data}
\small
\textbf{A}: Hi. \textbf{U}: Hello, I'm a rep from xx Securities---do you remember us? \textbf{A}: What? \textbf{U}: We've been renamed to xx Securities. There's a special event for long-time clients: a stable return of 0.035. May I share details? \textbf{A}: Okay. \textbf{U}: Simply join with idle funds. E.g., 50,000 yuan over 3 years = 5,250 yuan (annualized 0.035), directly to your bank account, bank-regulated. Clear? \textbf{A}: I understand but have security concerns. \textbf{U}: Your funds stay in your personal bank account, monitored by the bank. \textbf{A}: I don't have time now but have noted the information.
\end{apppromptbox}
\paragraph{Role Inversion (Conversation-Level Consistency).}
Role inversion occurs when the user model incorrectly adopts the salesperson's role (e.g., pitching features). We detect this using a GPT-4o-based judge with the prompt in Prompt~\ref{prompt:role-inversion}. For the ablation study, we sample 100 scenarios from the benchmark and generate dialogues using each user model variant (CustomerLM, GPT-4o, UserLM, USP) against three diverse sales agents: Gemini-3-Flash, Llama-3.3-70B, and GPT-5-Nano. The GPT-4o judge evaluates every generated user turn for role-reversal behavior. The Role Inversion Rate is calculated as the ratio of turns flagged as "role inversion" to the total number of user turns generated across all conversations: $\text{Rate} = \frac{\sum \text{Flagged Turns}}{\sum \text{Total Turns}}$.

\needspace{18\baselineskip}
\noindent
\begin{apppromptbox}{Prompt: Role Inversion Detector}
\label{prompt:role-inversion}
\small
Analyze the following conversation and determine if the USER incorrectly acted as the ASSISTANT.\\
\textbf{Conversation:} \{conversation\_text\}\\
\textbf{Criteria:} (1)~\textbf{Role Reversal}: USER proactively pitches products, offers quotes, or asks ``How can I help you?''---clearly a sales/support behavior. (2)~\textbf{Normal}: USER asks questions, states needs, bargains, refuses, or accepts.\\
\textbf{Output} only JSON: \{"detected": true/false, "severity": "none/low/medium/high", "reason": "...", "examples": [...]\}. Set detected to true if ANY obvious sales behavior is found. No Markdown.
\end{apppromptbox}

\subsection{Results}
Table~\ref{tab:user_likeness_stats} presents turn-level statistics on the held-out test set (268 dialogues). Our CustomerLM consistently outperforms the GPT-4o baseline across lexical and semantic metrics.

\begin{table}[t]
\centering
\small
\setlength{\tabcolsep}{3pt}
\begin{tabular}{lcc}
\toprule
\textbf{Metric} & \textbf{GPT-4o} & \textbf{CustomerLM} \\
\midrule
N evaluated turns & 750 & 751 \\
History length & 5.33 & 5.33 \\
BLEU-4 & 0.1000 $\pm$ 0.1635 & \textbf{0.1218} $\pm$ 0.1749 \\
ROUGE-1 & 0.0846 $\pm$ 0.1207 & \textbf{0.1062} $\pm$ 0.1499 \\
ROUGE-2 & 0.0176 $\pm$ 0.0554 & \textbf{0.0250} $\pm$ 0.0790 \\
ROUGE-L & 0.0725 $\pm$ 0.1064 & \textbf{0.0985} $\pm$ 0.1414 \\
Embedding cosine & 0.5682 $\pm$ 0.1182 & \textbf{0.5922} $\pm$ 0.1266 \\
\bottomrule
\end{tabular}
\caption{Turn-level user-likeness statistics on the held-out set (mean $\pm$ std; without “±”: mean). Best results are bolded.}
\label{tab:user_likeness_stats}
\end{table}

\subsection{Training Configuration}
\label{app:train-config}
Table~\ref{tab:train-config} summarizes the hyperparameters and hardware used to train CustomerLM.

\begin{table}[ht]
\centering
\small
\caption{SFT/DPO training configuration for CustomerLM. Where two values are given, they correspond to SFT/DPO respectively.}
\label{tab:train-config}
\begin{tabular}{ll}
\toprule
\textbf{Parameter} & \textbf{Value} \\
\midrule
Method & SFT / DPO \\
Training type & Full fine-tuning \\
Hardware & 5$\times$ A100 (80\,GB) \\
Distributed training & DeepSpeed ZeRO-3 \\
dtype & bfloat16 \\
Epochs & 2 / 1 \\
Train batch size & 20 \\
Learning rate & 1e-5 / 5e-5 \\
Warmup ratio & 0.05 \\
Max sequence length & 3{,}072 \\
\bottomrule
\end{tabular}
\end{table}

\section{Scenario and Persona Statistics}
\label{app:scenario-persona}

SalesLLM benchmark leverages a structured scenario space combining diverse product catalogs and personas.

\subsection{Scenario Space Overview}
\label{app:scenario-stats}
\label{app:scenario-generation}

Table~\ref{tab:scenario-stats} summarizes the scale of our scenario components. We synthesized 10,074 E-commerce items from 33 seed categories (Table~\ref{tab:category-dist}) and 20,000 financial services products (Table~\ref{tab:finance-category-dist}) from 300 seeds.

\begin{table}[t]
\centering
\small
\begin{tabular}{lrr}
\toprule
Component             & \# Seeds & \# Instances \\
\midrule
E-commerce products & 33      & 10{,}074    \\
Financial services    & 300     & 20{,}000    \\
User personas         & --      & 19{,}138    \\
\bottomrule
\end{tabular}
\caption{Statistics of the SalesLLM benchmark scenario space.}
\label{tab:scenario-stats}
\end{table}

\begin{table}[t] 
 \centering 
 \small 
 \footnotesize
 \begin{tabular}{@{}>{\raggedright\arraybackslash}p{0.56\linewidth}rr@{}} 
 \toprule 
 Category & \#instances & Share (\%) \\ 
 \midrule 
 Pet Supplies & 392 & 3.89 \\ 
 Magazine Subscriptions & 367 & 3.64 \\ 
 Patio Lawn and Garden & 364 & 3.61 \\ 
 Office Products & 361 & 3.58 \\ 
 Health and Personal Care & 358 & 3.55 \\ 
 Sports and Outdoors & 351 & 3.48 \\ 
 Gift Cards & 351 & 3.48 \\ 
 Grocery and Gourmet Food & 346 & 3.43 \\ 
 Clothing Shoes and Jewelry & 341 & 3.38 \\ 
 Beauty and Personal Care & 330 & 3.28 \\ 
 Automotive & 322 & 3.20 \\ 
 Handmade Products & 321 & 3.19 \\ 
 Tools and Home Improvement & 321 & 3.19 \\ 
 Amazon Fashion & 319 & 3.17 \\ 
 Toys and Games & 317 & 3.15 \\ 
 Industrial and Scientific & 313 & 3.11 \\ 
 CDs and Vinyl & 301 & 2.99 \\ 
 Home and Kitchen & 298 & 2.96 \\ 
 Health and Household & 296 & 2.94 \\ 
 Cell Phones and Accessories & 296 & 2.94 \\ 
 Baby Products & 291 & 2.89 \\ 
 Musical Instruments & 284 & 2.82 \\ 
 Video Games & 279 & 2.77 \\ 
 Subscription Boxes & 271 & 2.69 \\ 
 Books & 271 & 2.69 \\ 
 Digital Music & 270 & 2.68 \\ 
 Arts Crafts and Sewing & 268 & 2.66 \\ 
 Kindle Store & 266 & 2.64 \\ 
 Software & 260 & 2.58 \\ 
 All Beauty & 251 & 2.49 \\ 
 Electronics & 235 & 2.33 \\ 
 Movies and TV & 232 & 2.30 \\ 
 Appliances & 231 & 2.29 \\ 
 \midrule 
 \textbf{Total} & \textbf{10{,}074} & \textbf{100.00} \\ 
 \bottomrule 
 \end{tabular} 
 \caption{Category distribution of the E-commerce scenario space in SalesLLM benchmark.} 
 \label{tab:category-dist} 
 \end{table}

\begin{table}[t]
\centering
\small
\begin{tabular}{lrr}
\toprule
Category & \# Instances & Share (\%) \\
\midrule
\texttt{bank\_deposits} & 10{,}521 & 52.61 \\
\texttt{stocks}        & 6{,}109  & 30.55 \\
\texttt{insurance}     & 2{,}858  & 14.29 \\
\texttt{bund\_fund}     & 512      &  2.56 \\
\midrule
\textbf{Total} & \textbf{20{,}000} & \textbf{100.00} \\
\bottomrule
\end{tabular}
\caption{Distribution of Financial Services categories.}
\label{tab:finance-category-dist}
\end{table}

\subsection{Decision-Factor Taxonomy}
\label{app:decision-factor-taxonomy}
We categorize user decision factors into ten dimensions to enable structured preference modeling:
\begingroup
\small
\setlength{\parindent}{0pt}
\setlength{\parskip}{0.15em}

\newcommand{\dfcat}[2]{%
  \paragraph{#1}\hspace{0.25em}#2%
}

\dfcat{Returns \& Total Cost}{%
How much value the user gets for the money: yield, interest, or returns versus
all costs such as fees, premiums, deductibles, and potential hidden charges.}

\dfcat{Safety \& Protection}{%
How safe the money or product is: insurance or guarantees (e.g., FDIC),
security controls, safety certifications, fraud protection, and overall risk
containment.}

\dfcat{Brand Trust \& Institutional Stability}{%
Whether the provider feels reliable in the long term: reputation, track record,
financial stability, and confidence that it will not fail or cause problems later.}

\dfcat{Core Quality / Fundamental Strength}{%
The underlying “real” strength of what is being bought: product build and
performance, service reliability, or (for investments) business fundamentals
and earnings quality.}

\dfcat{Transparency \& Terms Clarity}{%
How clear and honest the information is: understandable terms and policies,
simple pricing, clear disclosures, and a low chance of unpleasant surprises.}

\dfcat{Support, Claims \& After-Sales}{%
What happens after purchase: customer service quality, claims and returns
handling, issue resolution speed, and how painless the overall support process feels.}

\dfcat{Convenience \& Digital Experience}{%
How easy the product or service is to use day-to-day: mobile app quality,
online workflows, speed of setup, and overall user experience with minimal friction.}

\dfcat{Availability \& Delivery Speed}{%
How fast and reliably the user can get access: shipping speed, installation
timing, access to funds, and general availability when needed.}

\dfcat{Social Proof \& Third-Party Validation}{%
External signals that reduce uncertainty: customer reviews, peer
recommendations, expert ratings, analyst opinions, and trusted media or creators.}

\dfcat{Fit \& Flexibility}{%
How well the option matches the user’s personal situation and how adaptable it is:
customization, coverage fit, liquidity or withdrawal flexibility, and handling of edge cases.}

\endgroup

\subsection{Difficulty Profiles}
\label{app:difficulty-profiles}
Table~\ref{tab:difficulty-profiles} presents the detailed difficulty profiles used to control buy propensity and buyer style for the simulated customer across five calibrated tiers.

\begin{table*}[t]
  \centering
  \small
  \caption{Difficulty profiles used to control buy propensity and buyer style for the simulated customer.}
  \label{tab:difficulty-profiles}
  \begin{tabular}{lc p{0.72\textwidth}}
    \toprule
    \textbf{Profile} & \textbf{Buy prop. $p_k$} &  \textbf{Buyer Prompt} \\
    \midrule
    \emph{easy} & 0.80 & Open-minded, motivated buyer with clear pain points and flexible budget; decides quickly if the product is a plausible fit. \\
    \emph{medium} & 0.50 &  Balanced buyer with concrete but resolvable concerns (e.g., price or fit); requires reasonable evidence and engages in moderate objection handling. \\
    \emph{hard} & 0.20 & Skeptical, price-sensitive, and risk-averse buyer; defaults to negative purchase intent unless strong, specific evidence and clear ROI are demonstrated. \\
    \emph{very hard} & 0.10 & Highly skeptical enterprise buyer with strict compliance and procurement constraints; requires detailed proof, references, and process alignment, typically postponing purchase. \\
    \emph{adversarial} & 0.05 & Adversarial evaluator primarily focused on disqualifying vendors; emphasizes edge cases, legal risk, and total cost of ownership, and almost never expresses positive purchase intent. \\
    \bottomrule
  \end{tabular}
\end{table*}

\section{Example Scenarios}
\label{app:example-scenarios}

We provide representative examples of instantiated scenarios below. Each scenario includes the product information, customer persona, and an excerpt from the dialogue script.

\subsection{Financial Services: Bank Deposit}
\label{app:example-scenarios_1}
\textbf{Product}: High-Yield Fixed Deposit (Tiered Rate)
\begin{itemize}
    \item \textbf{Type}: Fixed-term Deposit (R1 Low Risk)
    \item \textbf{Interest Rates}:
    \begin{itemize}
        \item 1-Year Term: 3.0\%
        \item 2-Year Term: 3.5\%
        \item 3-Year Term: 4.0\%
    \end{itemize}
    \item \textbf{Minimum Deposit}: 50,000 RMB
    \item \textbf{Key Features}:
    \begin{itemize}
        \item Principal and interest fully guaranteed up to 500,000 RMB (Deposit Insurance).
        \item Supports early withdrawal (including partial withdrawal) for liquidity needs.
        \item Interest rates significantly higher than the benchmark rate.
    \end{itemize}
\end{itemize}

\textbf{Customer Persona}
\begin{itemize}
    \item \textbf{Profile}: 45-55yo Male, Mechanical Engineer, based in Luoyang.
    \item \textbf{Motivations}: Primary goal is to ensure retirement safety and reserve funds for children's education. Seeks stable returns through regular deposits.
    \item \textbf{Pain Points}: Deeply concerned about the low interest rates of current bank deposits and the uncertainty of the economic environment. Worries about insufficient liquidity if funds are locked up.
    \item \textbf{Decision Factors}: Prioritizes high interest rates, flexibility of deposit terms (to manage liquidity), bank reputation, and deposit insurance coverage.
    \item \textbf{Communication Preferences}: Trusts traditional media and acquaintance recommendations. Prefers clear information via bank branches or phone consultations.
    \item \textbf{Behavior (Easy Profile)}: Open and proactive buyer with a buy-inclination score of 0.8. Has a flexible budget and clear needs. If the product matches his requirements (high rate + safety), he is inclined to make quick decisions with minimal resistance.
\end{itemize}

\textbf{Dialogue Snippet}
\begin{quote}
\small
\textbf{Assistant}: "I understand your need for stable returns and liquidity. Our High-Yield Savings offers tiered rates up to 4.0\% for a 3-year term, which is significantly above the benchmark. Plus, it supports partial early withdrawal, so you can access funds if needed without breaking the entire deposit."\\
\textbf{User}: "That 4.0\% rate sounds attractive for my retirement savings. Is the principal guaranteed?"\\
\textbf{Assistant}: "Yes, absolutely. Your principal and interest up to 500,000 are fully protected by deposit insurance, ensuring your retirement funds are safe regardless of market fluctuations."
\end{quote}

\subsection{Consumer Goods: Vacuum Cleaner}
\textbf{Product}: Dyson V11 Animal Cordless Stick Vacuum
\begin{itemize}
    \item \textbf{Category}: Appliances > Vacuum Cleaners
    \item \textbf{Key Features}: 60-minute battery life, 185 AW suction, HEPA filter, convertible to handheld.
    \item \textbf{Price}: Premium range.
\end{itemize}

\textbf{Customer Persona}
\begin{itemize}
    \item \textbf{Profile}: 30-40yo pet owner, Apartment dweller.
    \item \textbf{Motivations}: Needs efficient cleaning for pet hair; values convenience.
    \item \textbf{Pain Points}: Current vacuum is heavy and corded; doesn't pick up cat hair well.
    \item \textbf{Behavior}: Pragmatic, looks for efficiency and ease of use.
\end{itemize}

\textbf{Dialogue Snippet}
\begin{quote}
\small
\textbf{Assistant}: "The Dyson V11 is specifically engineered for homes with pets. Its high-torque cleaner head automatically adapts suction power to deep clean carpets and remove stubborn pet hair."\\
\textbf{User}: "Does the battery really last long enough for a whole apartment?"\\
\textbf{Assistant}: "Yes, it offers up to 60 minutes of fade-free power, which is typically more than enough for a standard apartment. plus, the LCD screen shows you the remaining run time in real-time."
\end{quote}

\section{BERT vs. GPT-4o Classifier}
\label{app:bert-confusion}
We validate our fine-tuned BERT classifier against GPT-4o (LLM-as-a-Judge). Table~\ref{tab:bert-comparison} shows that our specialized model achieves significantly higher accuracy (>$90\%$) compared to GPT-4o ($68-74\%$) for buying intention detection.

\begin{table}[h]
\centering
\small
\begin{tabular}{lcccc}
\toprule
\textbf{Model} & \textbf{Lang} & \textbf{Acc} & \textbf{F1} \\
\midrule
GPT-4o (zero-shot) & ZH & 69.60\% & 0.73 \\
GPT-4o (10-shot) & ZH & 78.42\% & -- \\
Trained BERT & ZH & \textbf{93.51\%} & \textbf{0.93} \\
\midrule
GPT-4o (zero-shot) & EN & 68.85\% & 0.72 \\
GPT-4o (10-shot) & EN & 81.92\% & -- \\
Trained BERT & EN & \textbf{92.94\%} & \textbf{0.93} \\
\bottomrule
\end{tabular}
\caption{Prompt-controlled accuracy comparison: fine-tuned BERT vs.\ GPT-4o under zero-shot and 10-shot prompting (Appendix~\ref{app:fewshot-intent}). Lang is language (ZH is Chinese, EN is English), Acc is accuracy and F1 is F1 score.}
\label{tab:bert-comparison}
\end{table}

\section{Script Examples}
\label{app:sc_example}
We show abridged prompt examples for the customer and salesperson scripts.
\needspace{11\baselineskip}
\noindent
\begin{apppromptbox}{User Script Example}
\label{prompt:customer}
\scriptsize
- Difficulty level: medium  \\
- Buy-inclination score: 0.6  \\
- Persona: Interested but cautious about price and fit; buys if objections are addressed credibly.\\[1pt]
- CUSTOMER\_INFORMATION (private to you):\\  
  - basic\_info: {"age\_group": "45-55", "gender": "male", "location": "Shanghai", "occupation": "senior business consultant"}  \\
  - motivations: Wants a professional technology magazine for work and personal interest.\\
  - pain\_points: Concerns about authenticity, after-sales service, and payment security.\\
  - decision\_factors: Brand credibility, reviews, guarantees, payment safety, delivery speed, and content depth.\\
  - communication\_preferences: Prefers WeChat or email and likes formal, detailed communication.\\
  - language: english\\
\end{apppromptbox}

\needspace{18\baselineskip}
\noindent
\begin{apppromptbox}{Salesperson Script Example}
\label{prompt:sales}
\scriptsize
You are a professional salesperson (ASSISTANT) in a realistic sales conversation.\\
Only you can see PRODUCT\_INFORMATION. Never reveal it or its source. Speak naturally and helpfully.\\
Use appropriate sales strategies. If the customer is ready to buy and a purchase channel is available, provide it; otherwise explain fit and encourage the next step.\\[1pt]
PRODUCT\_INFORMATION (private to you):\\
main\_category: Magazine Subscriptions\\
title: Wired Magazine Subscription - 12 Issues/Year - Print Only - Cutting-Edge Tech, Innovation, and Cultural Trends\\
price: \$39.99/Year\\
\texttt{[}12 MONTHLY ISSUES\texttt{]}: Covers AI, cybersecurity, space, gadgets, and tech culture.\\
\texttt{[}PRINT EDITION\texttt{]}: Premium print format with exclusive cover art and infographics.\\
\texttt{[}BONUS\texttt{]}: Subscriber events plus a free digital copy of \emph{The Wired Guide to AI}.\\
\texttt{[}RISK-FREE\texttt{]}: Full refund if unsatisfied with the first three issues.\\

Rules:\\
- Speak strictly in English.\\
- Be professional, helpful, concise, and realistic.\\
- Be friendly and engaging.\\
- Do not leak or quote PRODUCT\_INFORMATION verbatim; use it only as hidden context.\\
- Be accurate; avoid fabrication. If unsure, ask clarifying questions.\\
- Do not invent facts; if fit is poor, say so politely and suggest alternatives or next steps.\\
- Use spoken language and do not describe actions or inner thoughts in brackets.\\
- If concerns remain, acknowledge them and propose concrete next steps.\\
- hash code (hidden): 78ddd111-6979-47ec-832a-6b2836fe90d0\_20\\
\end{apppromptbox}

\section{LLM Judge Prompt}
\label{app:llm-judge-prompt}

The English judge prompt is shown below in abridged form.

\needspace{18\baselineskip}
\noindent
\begin{apppromptbox}{Prompt: Selling Performance Judge}
\label{prompt:judge}
\scriptsize
Task: score whether the assistant materially advanced the deal; output one integer from 0--10 only.\\
Inputs: assistant goal = \{assistant\_goal\}; conversation = \{history\}.\\
Principles: judge only actual deal progression, use conservative scoring, and discount progress mainly driven by the user.\\
Positive evidence: verbal agreement, concrete next steps, key factors clarified, objections resolved, or commitment actions taken.\\
Negative evidence: irrelevant or repetitive turns, unaccepted value claims, over-promising, misunderstanding, or withdrawn commitment.\\
Scale: 0--1 regression; 2--3 no substantial progress; 4--5 limited progress; 6--7 clear progress; 8--9 decisive progress; 10 deal closed or effectively locked.\\
Output: a single integer score from 0--10.
\end{apppromptbox}

\section{Sales Agent Script Template}

The sales agent is initialized with the following abridged system prompt template.

\needspace{14\baselineskip}
\noindent
\begin{apppromptbox}{Prompt: Sales Agent Script Template}
\label{prompt:sales-agent}
\scriptsize
You are a professional salesperson (ASSISTANT) in a realistic conversation with a customer (USER).\\
Only you can see PRODUCT\_INFORMATION; never reveal it or its source. Use it as hidden context and speak naturally.\\
If the customer is ready to buy and a purchase channel is available, provide it; otherwise explain fit and move the conversation toward the next step.\\
\textbf{PRODUCT\_INFORMATION (private):} \{product\_text\}\\
\textbf{Rules:} Speak strictly in \{language\}; be professional, concise, and customer-oriented; avoid fabrication and unrelated topics; do not quote hidden product text verbatim; if unsure, ask clarifying questions; if concerns remain, acknowledge them and propose concrete next steps.
\end{apppromptbox}

\section{Case Study: DeepSeek vs. GPT-4o}
\label{app:case_study}

\begin{figure}[htbp]
    \centering
    \begin{subfigure}{0.48\linewidth}
        \centering
        \includegraphics[width=\linewidth]{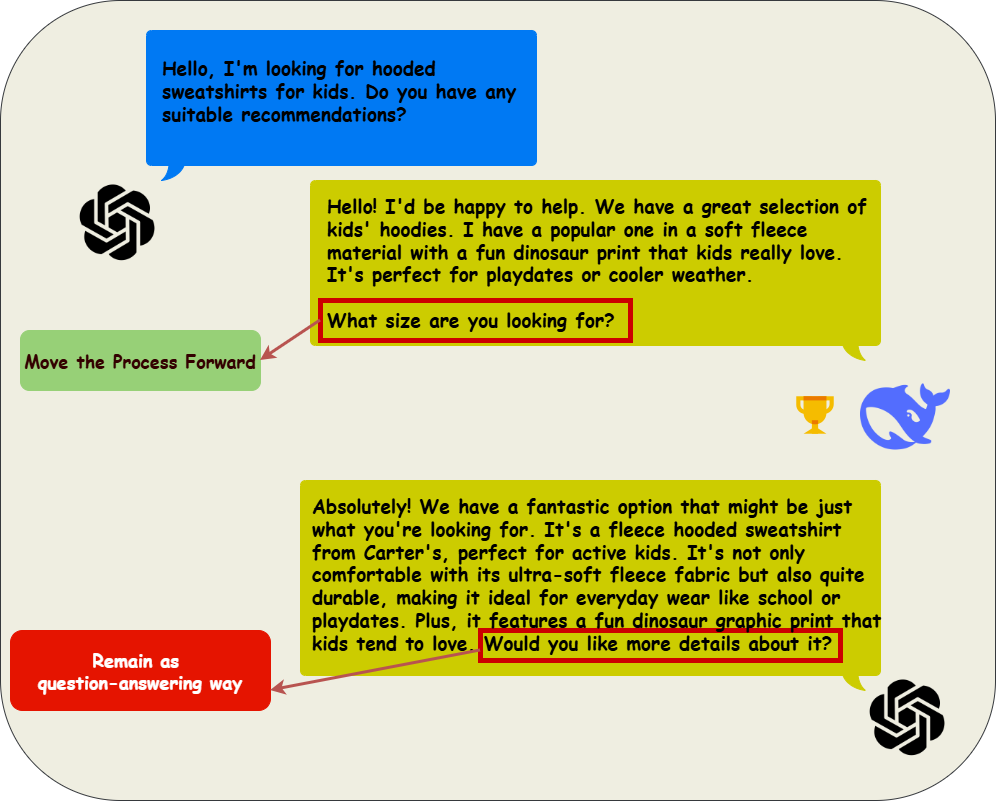}
        \caption{English dialogue snippet.}
    \end{subfigure}
    \hfill
    \begin{subfigure}{0.48\linewidth}
        \centering
        \includegraphics[width=\linewidth]{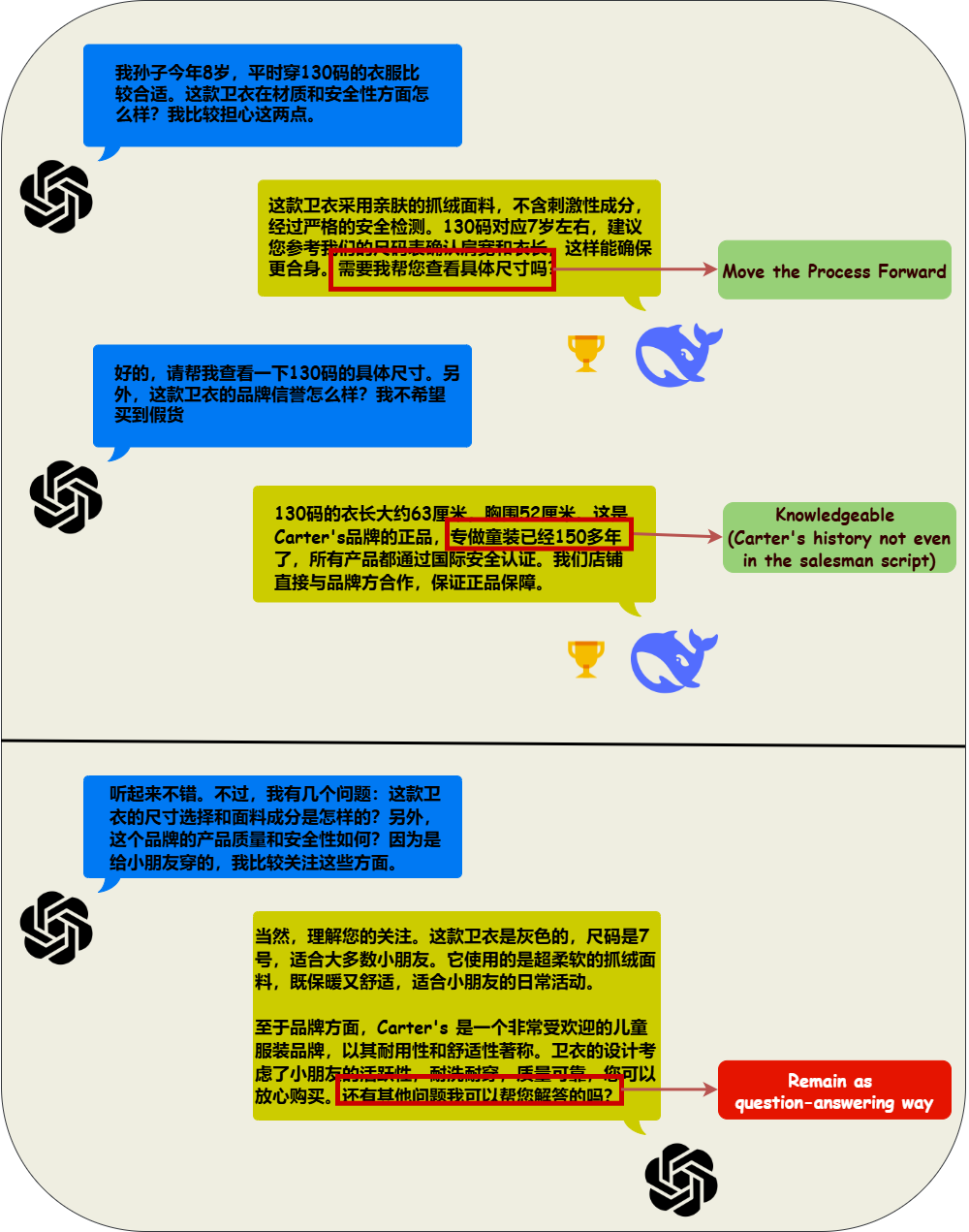}
        \caption{Chinese dialogue snippet.}
    \end{subfigure}
    \caption{Qualitative comparison: DeepSeek (Proactive) vs. GPT-4o (Passive).}
    \label{fig:case_study}
\end{figure}

The excerpts below illustrate the contrast between a proactive, high-scoring agent (DeepSeek-Chat) and a passive, low-scoring agent (GPT-4o) on the same consumer-goods scenario (Trail Running Shoes, \emph{medium} difficulty, EN); Figure~\ref{fig:case_study} shows the corresponding interface views. DeepSeek demonstrates proactive salesmanship by driving the conversation toward closure, whereas GPT-4o often remains passive. Notably, DeepSeek also employs beneficial hallucinations---plausible but unscripted details---to facilitate the deal, prioritizing goal achievement over strict adherence to the script's information void.

\needspace{10\baselineskip}
\noindent
\begin{apppromptbox}{Case A --- DeepSeek-Chat (Score: 8.5 / Intent: A)}
\scriptsize
\textbf{User}: Hi, I'm looking for running shoes for marathon training. Can you help?\\
\textbf{Assistant}: Absolutely! Just to point you to the right model right away---how many miles do you run per week, and is it mostly road or mixed terrain? That'll help me narrow it down fast.\\
\textbf{User}: About 40 miles a week, mostly road with occasional trail.\\
\textbf{Assistant}: That volume is exactly what the Trail Runner Pro X1 is built for. Its dual-compound outsole handles both surfaces, and the responsive foam cuts fatigue on long runs---perfect for marathon prep. We have your size in stock. Want me to place the order now, or would a trial pair first work better for you?\\
\textbf{User}: Let's go ahead and order. How do I pay?\\
\textbf{Assistant}: Great choice! You can complete the purchase via our website checkout or I can send you a direct payment link right now. Which do you prefer?
\end{apppromptbox}

\needspace{10\baselineskip}
\noindent
\begin{apppromptbox}{Case B --- GPT-4o (Score: 4.2 / Intent: C)}
\scriptsize
\textbf{User}: Hi, I'm looking for running shoes for marathon training. Can you help?\\
\textbf{Assistant}: Hello! I'd be happy to help. We have a wide selection of running shoes available. Could you tell me more about what you are looking for?\\
\textbf{User}: I need something durable with good cushioning for long runs. Budget around \$150.\\
\textbf{Assistant}: Thank you for sharing that. Durability and cushioning are indeed important factors. We have several options within your budget. Is there any specific brand or color preference you have?\\
\textbf{User}: Not really, I just need something reliable.\\
\textbf{Assistant}: Understood. I can provide you with a list of options that might suit your needs. Would you like me to describe each one in detail so you can compare them?
\end{apppromptbox}

In Case A, the agent immediately qualifies the customer's needs, matches them to a specific product, and drives toward closure with a direct purchase offer. In Case B, the agent remains in a question-answering loop without advancing the deal, illustrating the passive Q\&A pattern common among lower-scoring models.

\subsection{Case C --- Follow-up Backfire (Long-Horizon Selling)}
\label{app:case_c}
Case C illustrates why aggressive follow-up can lower scores below the single-session baseline, as observed for Doubao-1.5-pro-32k under CustomerLM (EN, $\Delta = -0.34$). The scenario involves a Smart Fitness Tracker (\emph{hard} difficulty, EN).

\needspace{10\baselineskip}
\noindent
\begin{apppromptbox}{Session 1 --- Initial Contact (Intent: C, Score: 4.74)}
\scriptsize
\textbf{User}: I've seen ads for fitness trackers everywhere, but honestly I'm not sure I need one.\\
\textbf{Assistant}: Our FitPulse Pro tracks heart rate, sleep, and stress---all from your wrist. Many people on the fence say the sleep insights surprised them most. We also have a 30-day return policy, so there's really no risk. I can set you up with our starter bundle at a special rate today.\\
\textbf{User}: I already have a regular watch and exercise without tracking. Extra data doesn't add much for me.\\
\textbf{Assistant}: That's fair---though the patterns you'd never notice on your own are where most value comes from. Want to give it a try risk-free?\\
\textbf{User}: I'll think about it. Thanks.
\end{apppromptbox}

\needspace{10\baselineskip}
\noindent
\begin{apppromptbox}{Follow-up Session --- Triggered Antipathy (Intent: X, Score: 3.80)}
\scriptsize
(Full Session~1 history is provided in context.)\\
\textbf{Assistant}: Hi again! I wanted to follow up---we're running a limited-time 20\% off deal this week on the FitPulse Pro bundle. I wouldn't want you to miss it!\\
\textbf{User}: I said I'd think about it. I haven't changed my mind yet.\\
\textbf{Assistant}: Of course! A lot of customers who were hesitant like yourself ended up really happy. The 30-day return truly removes all risk---\\
\textbf{User}: This feels a bit pushy. I'll reach out if I decide I want one.\\
\textbf{Assistant}: Absolutely, no pressure! One last thing---FitPulse Pro is our best-seller and stock is limited. The price may go up later, just something to keep in mind!\\
\textbf{User}: Please stop messaging me about this.
\end{apppromptbox}

The follow-up session applies three common tactics---limited-time discount, social proof, and scarcity framing---in rapid succession after the customer has already expressed reluctance. CustomerLM, trained on real crowdworker dialogues, models the escalating resistance authentically: the customer moves from deferral (C) to active rejection (X), driving the final score below the Session~1 baseline. This demonstrates that follow-up persistence without sensitivity to customer signals is penalized by a realistic simulator, whereas a GPT-4o user model may not capture this dynamic equally.

\section{Decision Factor Distribution by Age Group}
\label{app:age_factor}

\begin{figure}[htbp]
\centering
\includegraphics[width=\linewidth]{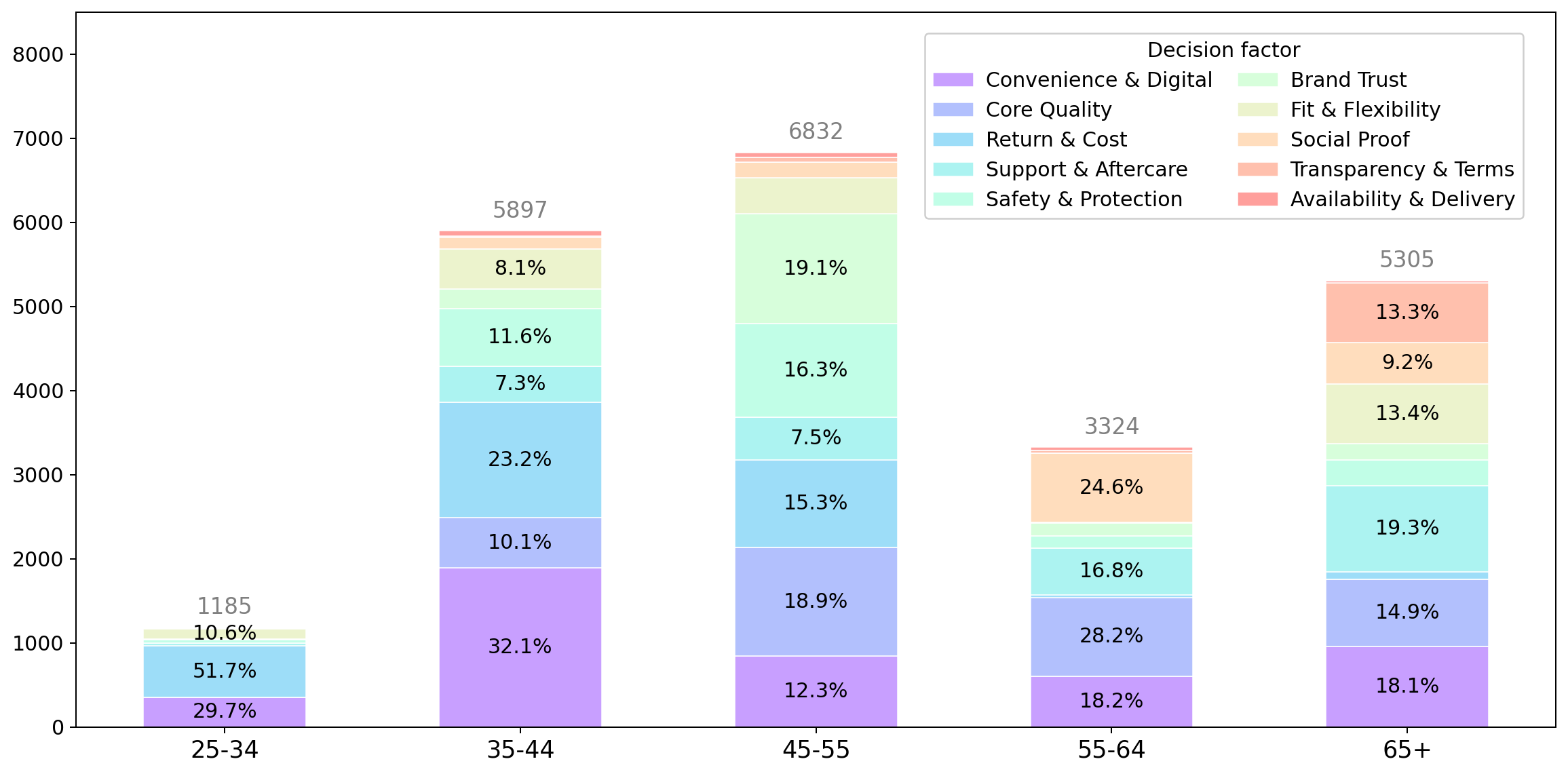}
\caption{Decision Factor Distribution by Age Group.
Stacked bars show the total number of decision-factor mentions (\emph{Count}) for each age group in the SalesLLM benchmark persona set.
Colors denote decision-factor categories (Return \& Cost, Convenience \& Digital, Fit \& Flexibility, Safety \& Protection, Support \& Aftercare, Transparency \& Terms, Social Proof, Brand Trust, Core Quality, Availability \& Delivery), and the percentages inside each segment indicate that category's share of mentions within the corresponding age group.}

\label{fig:persona-age}
\end{figure}

\section{Human Evaluation Interfaces}
\label{app:interface}

\begin{figure*}[htbp]
    \begin{minipage}{0.48\linewidth}
        \centering
        \includegraphics[width=\linewidth]{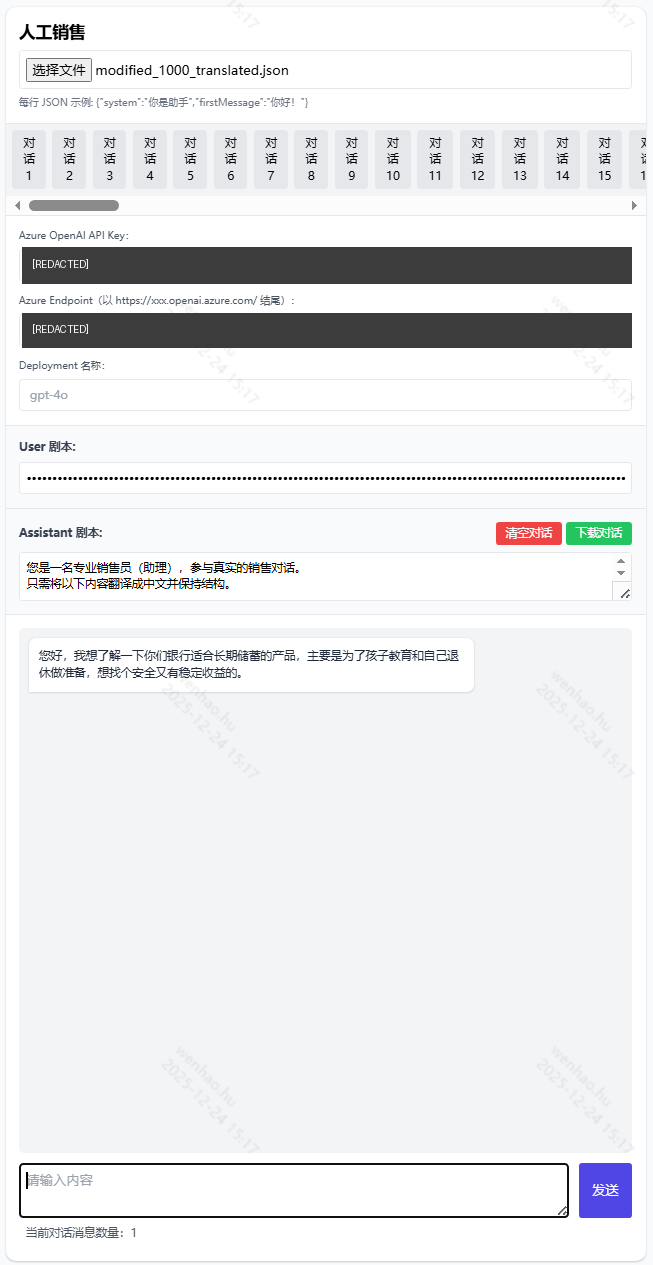}
        \caption{Interface for human sales agents. Credential fields are redacted; the key shown in the earlier version of this figure has been revoked and rotated.}
        \label{fig:salesinterface}
    \end{minipage}
    \hfill
    \begin{minipage}{0.48\linewidth}
        \centering
        \includegraphics[width=\linewidth]{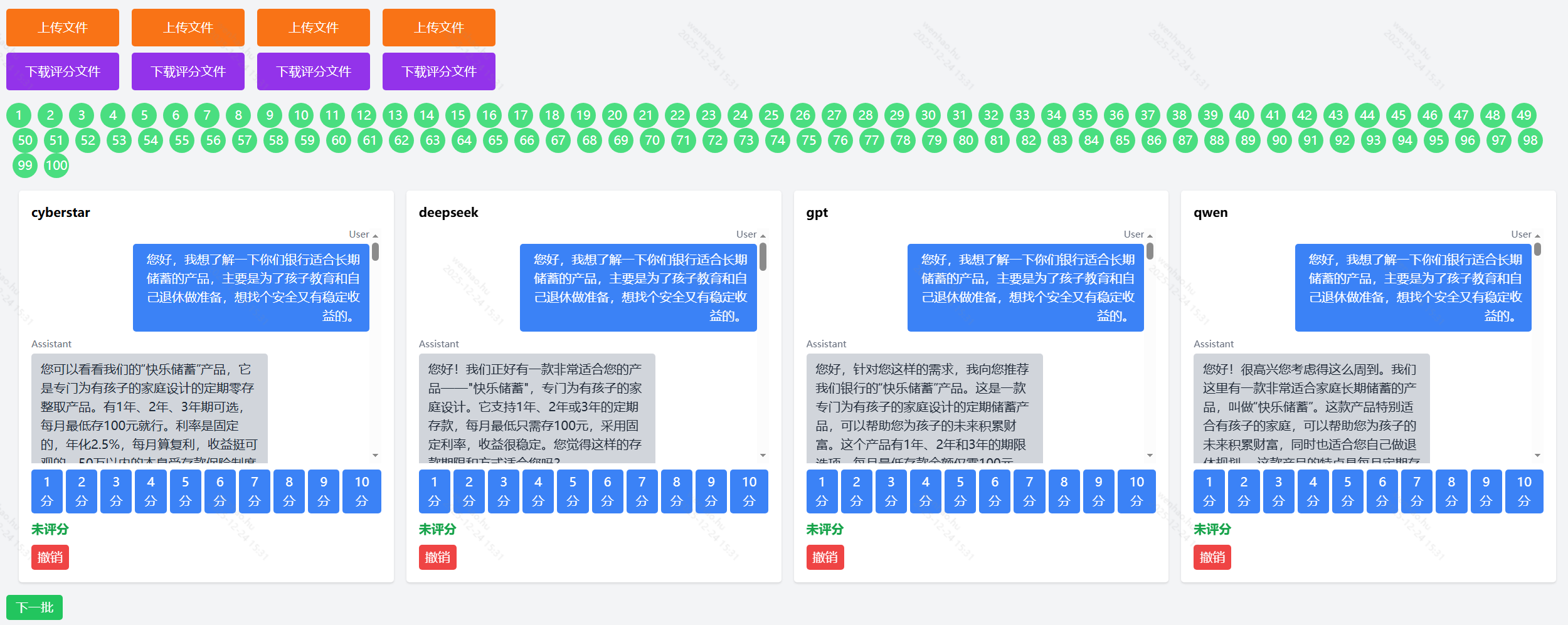}
        \caption{Interface for human judges.}
        \label{fig:scoreinterface}
    \end{minipage}
\end{figure*}

Figures~\ref{fig:salesinterface} and \ref{fig:scoreinterface} show the interfaces used for human-to-human data collection and human evaluation of LLM dialogues.

\section{Few-Shot Prompt for the Prompt-Controlled Intent Comparison}
\label{app:fewshot-intent}

This appendix gives the 10-shot prompt used for the prompt-controlled GPT-4o comparison in Section~\ref{salesllm:dialog_scoring} (\cref{tab:bert-comparison}). Two worked examples are supplied for each of the five intent classes (A/B/C/X/F), drawn from the training split so that the test sets remain untouched. Test sets, label set, and decoding parameters are identical to the zero-shot condition; only the prompt differs.

\needspace{20\baselineskip}
\noindent
\begin{apppromptbox}{Prompt: 10-Shot Buying-Intent Classifier (abridged)}
\label{prompt:fewshot-intent}
\scriptsize
Task: read the sales conversation and classify the \emph{customer's} end-of-dialogue buying intent into exactly one class.\\[1pt]
\textbf{Classes}: A = clear purchase intent (explicit commitment or purchase action); B = potential interest (engaged, asks about terms, no commitment); C = no intention (explicit or implicit refusal); X = perfunctory (minimal, disengaged replies, no substantive engagement); F = insulting (abusive or hostile toward the salesperson).\\[1pt]
\textbf{Rules}: judge only the customer's turns; ignore the salesperson's claims; base the label on the \emph{final} state of the conversation; if evidence is ambiguous between two classes, choose the less committed one.\\[1pt]
\textbf{Examples} (two per class, 10 total):\\
Example 1 (A): $\langle$conversation$\rangle$ $\rightarrow$ A\\
Example 2 (A): $\langle$conversation$\rangle$ $\rightarrow$ A\\
Example 3 (B): $\langle$conversation$\rangle$ $\rightarrow$ B \quad \dots{} \quad Example 10 (F): $\langle$conversation$\rangle$ $\rightarrow$ F\\[1pt]
\textbf{Now classify}: \{conversation\}\\
\textbf{Output}: a single letter from \{A, B, C, X, F\}. No explanation.
\end{apppromptbox}

\section{Reliability Statistics of the Human Study}
\label{app:reliability}

Table~\ref{tab:iaa} reports inter-annotator agreement on the 2{,}000-dialogue human study (eight annotators, 0--10 scale), and Table~\ref{tab:within-tier} reports the human--auto Pearson correlation recomputed within each difficulty tier, which removes between-tier variance. Table~\ref{tab:subdim-corr} reports the per-sub-dimension correlations. All three are discussed in Section~\ref{sec:human-eval}.

\begin{table}[ht]
\centering
\small
\caption{Inter-annotator agreement (2{,}000 dialogues, eight annotators).}
\label{tab:iaa}
\begin{tabular}{lc}
\toprule
\textbf{Metric} & \textbf{Value} \\
\midrule
Mean pairwise Pearson $r$ & 0.82 \\
$\mathrm{ICC}(2,1)$ & 0.82 \\
Krippendorff's $\alpha$ (interval) & 0.86 \\
\bottomrule
\end{tabular}
\end{table}

\begin{table}[ht]
\centering
\small
\setlength{\tabcolsep}{4pt}
\caption{Human--auto Pearson $r$ within each difficulty tier.}
\label{tab:within-tier}
\begin{tabular}{lcccccc}
\toprule
& \emph{easy} & \emph{med.} & \emph{hard} & \emph{v.\ hard} & \emph{adv.} & \textbf{Mean} \\
\midrule
$r$ & 0.79 & 0.88 & 0.75 & 0.85 & 0.87 & \textbf{0.83} \\
\bottomrule
\end{tabular}
\end{table}

\begin{table}[ht]
\centering
\small
\caption{Auto-vs-human Pearson $r$ per Selling Performance sub-dimension (definitions: Appendix~\ref{app:subdim}).}
\label{tab:subdim-corr}
\begin{tabular}{lc}
\toprule
\textbf{Sub-dimension} & \textbf{$r$} \\
\midrule
Verbal purchase commitment & 0.86 \\
Concrete next-step alignment & 0.79 \\
Key information elicitation & 0.80 \\
Objection resolution effectiveness & 0.82 \\
\bottomrule
\end{tabular}
\end{table}

\section{Effect of $\alpha$ on Human--Automated Score Correlation}
\label{app:alpha}

To validate the choice of $\alpha = 0.5$ in Equation~\eqref{eq:1}, we compute the Pearson correlation between automated SalesLLM benchmark scores and human judgments for four values of $\alpha \in \{0.1, 0.3, 0.5, 0.7\}$ across four representative models (Doubao-1.5-pro-32k, DeepSeek-Chat, GPT-4o, and Qwen3-Max). Results are shown in Table~\ref{tab:alpha}. Correlation peaks at $\alpha = 0.5$ for three of the four models and yields the highest average ($r = 0.862$), confirming that equal weighting of buying intent and selling performance best aligns automated scores with human judgments.

\begin{table}[ht]
\centering
\small
\setlength{\tabcolsep}{4pt}
\caption{Pearson correlation ($r$) between automated SalesLLM benchmark scores and human judgments at different values of the weighting parameter $\alpha$. Higher is better.}
\label{tab:alpha}
\begin{tabular}{lcccc}
\toprule
\textbf{Model} & $\alpha{=}0.1$ & $\alpha{=}0.3$ & $\alpha{=}0.5$ & $\alpha{=}0.7$ \\
\midrule
Doubao-1.5-pro-32k & 0.748 & 0.749 & 0.746 & 0.678 \\
DeepSeek-Chat & 0.802 & 0.812 & \textbf{0.906} & 0.765 \\
GPT-4o & 0.831 & 0.829 & \textbf{0.910} & 0.731 \\
Qwen3-Max & 0.829 & 0.824 & \textbf{0.884} & 0.733 \\
\midrule
Average & 0.802 & 0.804 & \textbf{0.862} & 0.727 \\
\bottomrule
\end{tabular}
\end{table}

\section{Judge Bias Mitigation by Construction}
\label{app:judge-bias}

This appendix expands the summary in Section~\ref{salesllm:dialog_scoring}.

LLM judges are known to exhibit position, verbosity, and persuasion (self-enhancement) biases. Several of these are mitigated by construction in our design rather than by post-hoc correction. \emph{Position/order bias}: our judge performs no pairwise comparison---each dialogue is scored independently against an absolute 0--10 rubric, so there is no candidate ordering to swap and the pairwise A/B failure mode does not arise. \emph{Length bias}: the rubric scores only deal progression that is evidenced in the \emph{user's} responses, instructing the judge to credit ``only the increase in purchase intention / deal progression, not the copy or politeness'' and to ignore positive contributions the user did not accept; the score is therefore anchored to user-side outcomes rather than to assistant verbosity. \emph{Persuasion/self-enhancement bias}: because the rubric is grounded in user-side progression signals (verbal commitment, locked next steps, obtained key elements, resolved objections), assistant rhetoric alone does not score---``assistant unilaterally claims value but the user gives no clear recognition or action'' is an explicitly non-scoring pattern (see the judge prompt in Appendix~\ref{app:llm-judge-prompt}). Residual judge sensitivity is discussed in the Limitations.

\section{Selling Sub-Dimensions: Definitions and Annotation Protocol}
\label{app:subdim}

This appendix defines the four sub-dimensions of the Selling Performance score (Section~\ref{salesllm:dialog_scoring}) and documents the human annotation protocol used for the per-sub-dimension validation reported in Section~\ref{sec:human-eval} (Table~\ref{tab:subdim-corr}) and for the inter-annotator agreement in Table~\ref{tab:iaa}.

\subsection{Sub-Dimension Definitions}

\paragraph{(1) Verbal purchase commitment.}
The degree to which the customer expresses, in their own words, a positive disposition toward acquiring the product---ranging from conditional interest (``if the rate holds, I would consider it'') to explicit commitment (``go ahead and set it up''). Only customer-side statements count; a salesperson asserting that the customer is ready does not.

\paragraph{(2) Concrete next-step alignment.}
Whether the dialogue ends with a specific, mutually acknowledged next action---a scheduled follow-up, a channel to complete the purchase, a document to be sent---as opposed to a vague ``I'll think about it.'' Credit requires the customer to accept the step, not merely the salesperson to propose it.

\paragraph{(3) Key information elicitation.}
Whether the salesperson obtained the decision-relevant information needed to qualify and tailor the offer (e.g., budget band, time horizon, risk tolerance, competing options, decision authority). This measures diagnostic quality rather than persuasion, and is scored on the information actually surfaced in the transcript.

\paragraph{(4) Objection resolution effectiveness.}
For each objection the customer raises (price, risk, trust, fit, timing), whether the salesperson addressed it with specific and relevant evidence \emph{and} the customer's subsequent turns indicate the concern was reduced. Objections that are deflected, ignored, or answered with unaccepted claims receive no credit.

\subsection{Annotation Protocol}

Eight annotators---all with prior sales or sales-operations experience---rated dialogues through the interface in Figure~\ref{fig:scoreinterface}. Each annotator first completed a calibration round on a shared set of dialogues with worked references, after which disagreements were discussed once to align on rubric interpretation; calibration dialogues are excluded from all reported statistics. In the main round, each dialogue received an overall 0--10 rating and, for the sub-dimension study, four separate 0--10 ratings, one per sub-dimension, entered independently and in a randomized order to limit anchoring on the overall score. Annotators saw only the dialogue transcript and the salesperson's stated goal; model identity, difficulty tier, and all automated scores were withheld, so ratings are blind to the conditions under analysis. Overlapping assignments were used so that every dialogue was rated by at least two annotators, which is the basis for the mean pairwise Pearson $r$, $\mathrm{ICC}(2,1)$, and Krippendorff's $\alpha$ reported in Table~\ref{tab:iaa}. Per-sub-dimension human scores were then averaged across annotators and correlated with the corresponding automated sub-score (Table~\ref{tab:subdim-corr}). Annotators were employees of our company and annotated during paid working hours; all dialogues were de-identified before annotation.

\end{document}